\pdfoutput=1
\documentclass{article}

\PassOptionsToPackage{sort&compress}{natbib}
\usepackage[final]{corl_2026}
\makeatletter
\renewcommand{\@conferencelocation}{Austin, Texas, USA}
\makeatother
\usepackage{graphicx}
\usepackage{float}
\usepackage{wrapfig}
\usepackage{amsmath}
\usepackage{amssymb}
\usepackage{booktabs}
\usepackage{colortbl}
\usepackage[normalem]{ulem}
\usepackage{subcaption}

\definecolor{humiFigRed}{rgb}{0.604,0.000,0.004}
\definecolor{humiFigGreen}{rgb}{0.149,0.325,0.086}
\definecolor{humiTableGray}{gray}{0.92}

\title{HuMI: Humanoid Whole-Body Manipulation from Humanoid-Free Demonstrations}

\author{%
    Ruiqian Nai$^{1,2,3,*}$, Boyuan Zheng$^{1,2,3,*}$, Junming Zhao$^{1,2,3,*}$,\\
    \bfseries Haodong Zhu$^{1}$, Sicong Dai$^{1,\dagger}$, Zunhao Chen$^{1}$, Yihang Hu$^{1,2}$,\\
    \bfseries Yingdong Hu$^{1,2}$, Tong Zhang$^{1,2}$, Chuan Wen$^{4}$, Yang Gao$^{1,2,3,\ddagger}$\\[8pt]
    \normalfont $^{1}$IIIS, Tsinghua University \quad $^{2}$Shanghai Qi Zhi Institute\\
    \normalfont $^{3}$Spirit.AI \quad $^{4}$Shanghai Jiao Tong University
}

\hypersetup{
    pdftitle={HuMI: Humanoid Whole-Body Manipulation from Humanoid-Free Demonstrations},
    pdfauthor={Ruiqian Nai, Boyuan Zheng, Junming Zhao, Haodong Zhu, Sicong Dai, Zunhao Chen, Yihang Hu, Yingdong Hu, Tong Zhang, Chuan Wen, Yang Gao}
}

\begin{document}
\maketitle

\begingroup
\renewcommand{\thefootnote}{\fnsymbol{footnote}}
\footnotetext[1]{Equal contribution.}
\footnotetext[2]{Work done during an internship at Tsinghua University.}
\footnotetext[3]{Corresponding author.}
\endgroup

\begin{abstract}
Current approaches for humanoid visuomotor whole-body manipulation, primarily relying on teleoperation or visual sim-to-real reinforcement learning, are hindered by hardware logistics and complex reward engineering. Consequently, demonstrated autonomous skills remain limited and are typically restricted to controlled environments. In this paper, we present the Humanoid Manipulation Interface (HuMI), a framework for learning whole-body manipulation tasks across various environments. HuMI enables humanoid-free data collection by capturing rich whole-body motion using portable hardware. This data drives a hierarchical learning pipeline that translates human motions into dexterous and feasible humanoid skills. Extensive experiments across five whole-body tasks, including kneeling, squatting, tossing, walking, and bimanual manipulation, demonstrate that HuMI achieves a 3$\times$ increase in data collection efficiency compared to teleoperation and attains a 70\% success rate in unseen environments.
Project website: \url{https://humanoid-manipulation-interface.github.io/}.
\end{abstract}

\keywords{humanoid whole-body manipulation, data acquisition, robot learning}

\begin{figure}[H]
    \centering
    \includegraphics[width=\linewidth]{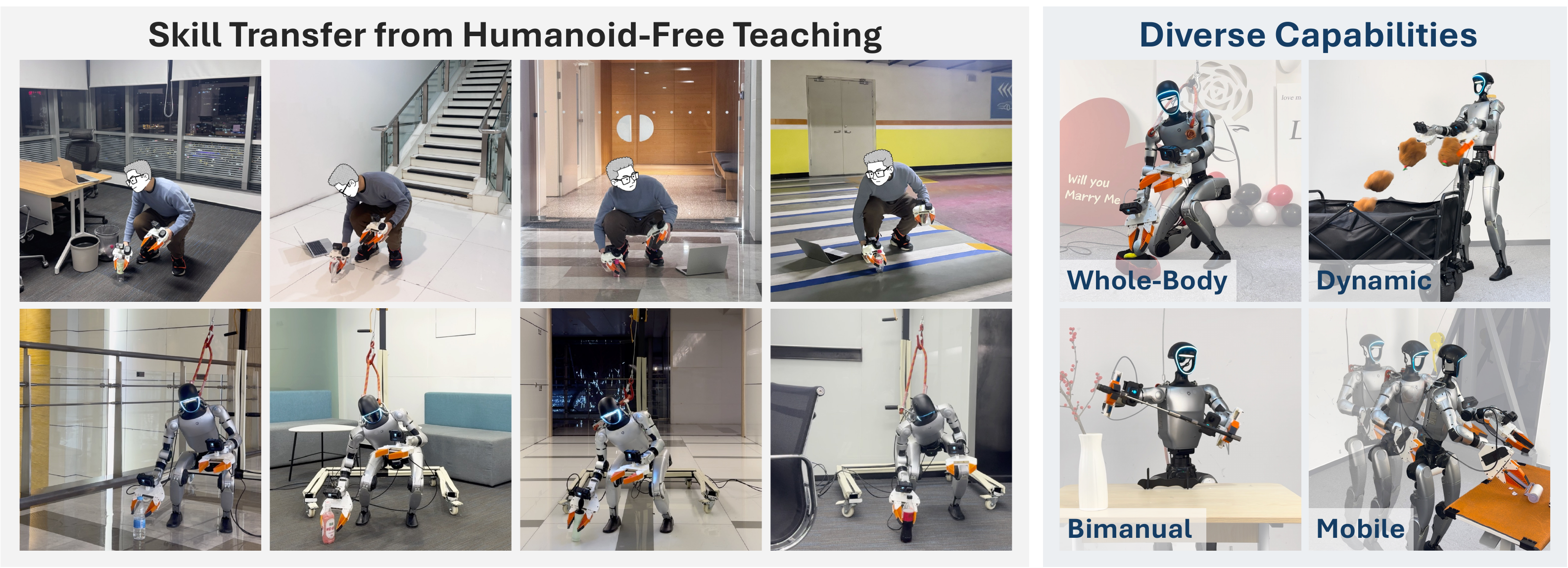}
    \caption{\textbf{Humanoid Manipulation Interface (HuMI)}. \textbf{Left:} A portable, humanoid-free data collection setup for transferring skills from humans to humanoids across varied environments. \textbf{Right:} Whole-body behaviors learned with the framework.}
    \label{fig:teaser}
\end{figure}

\section{Introduction}
\label{sec:intro}
Humans expertly coordinate their entire bodies for everyday manipulation---squatting to retrieve a bottle on the ground, or bending to wipe a low table. Enabling humanoid robots to acquire such visuomotor whole-body skills has largely relied on sim-to-real reinforcement learning (RL) \cite{xue2025opening,he2025viral,lin2025sim,luo2025emergent} or imitation learning from teleoperation \cite{ben2025homie,li2025clone,ze2025twist,ze2025twist2,li2025amo,fu2024humanplus,he2024omnih2o,ze2025generalizable,cheng2024open,cheng2024expressive,jiang2025wholebodyvla,luo2025sonic,liu2025trajbooster,wei2026psi_0}. These approaches incur substantial overhead: RL requires intricate reward shaping and domain randomization, while teleoperation requires operators to maintain the robot's balance and compensate for tracking errors. Consequently, scaling to diverse tasks and environments remains difficult, and existing systems have typically been shown on only a limited set of tasks in fixed laboratory settings.

In this work, our goal is to enable humanoid robots to perform diverse tasks across many environments, with an emphasis on whole-body coordination that fully exploits the dexterity of humanoid platforms. To this end, we propose the \textbf{Hu}manoid \textbf{M}anipulation \textbf{I}nterface (\textbf{HuMI}) (Fig. \ref{fig:teaser}), a \textbf{humanoid-free, portable, and efficient} data-collection and learning framework that requires only handheld sensorized grippers and base-station-free wearable pose trackers that fit into a single backpack. Achieving these capabilities, however, demands more than directly applying existing teleoperation-free data-collection systems \cite{chi2024universal,ha2024umi,gupta2025umi}. Such systems typically decompose into three parts: a recorder captures end-effector demonstrations, a high-level policy maps onboard observations to target end-effector trajectories, and a low-level controller tracks them. This pipeline is effective on tabletop arms but insufficient for humanoid whole-body manipulation.

First, demonstrations are \textbf{underspecified}: end-effector trajectories alone leave whole-body degrees of freedom ambiguous. We therefore also record pelvis and foot trajectories, and use inverse kinematics (IK) to recover the full robot configuration. Second, a \textbf{feasibility gap} arises because morphological discrepancies render human demonstrations kinematically infeasible, while naive motion scaling \cite{yang2025omniretarget,Luo2023PerpetualHC,joao2025gmr} is not viable, since scaling the body trajectory misaligns it with objects in the scene. We address this from a system-design perspective, providing demonstrators with an online IK preview that visualizes the retargeted motion in real time so they can adjust on the fly. Third, despite advances in humanoid tracking \cite{zhang2025hub, zhao2025resmimic, yang2025omniretarget, liao2025beyondmimic, he2025hover, su2025hitter, luo2025sonic, ben2025homie, ze2025twist2}, \textbf{non-negligible execution errors} (4--6\,cm) persist and break standard policy interfaces. We bridge this with a manipulation-centric whole-body controller and a redesigned policy interface for tighter high- and low-level coordination.

We evaluate HuMI on five tasks: single-knee kneeling pickup, squatting to retrieve a bottle from the ground, tossing a toy, unsheathing a sword, and walking to clean a table. These tasks span a variety of whole-body manipulation behaviors. We further evaluate generalization to unseen environments and objects, achieving a 70\% success rate. In summary, our contributions are:
\begin{itemize}
    \item A humanoid-free demonstration system for humanoid whole-body manipulation tasks.
    \item A hierarchical learning framework that transfers manipulation skills from humans to humanoids, with coordinated designs across controller training and the policy interface.
    \item Real-world validation on five whole-body tasks, with a 3$\times$ data-collection throughput over teleoperation and a 70\% success rate in unseen environments.
\end{itemize}

\section{Related Works}
\textbf{Humanoid whole-body manipulation.} Prior work can be broadly grouped by the state information assumed at deployment: methods that require access to ground-truth object poses \cite{wang2023physhoi,wang2025physhsi,weng2025hdmi,zhao2025resmimic,lin2026pro,lin2026lessmimic}, and visuomotor methods that rely solely on onboard sensors and image observations. We target the visuomotor setting, as it is the regime required for real-world deployment. Existing visuomotor approaches predominantly resort to either sim-to-real RL \cite{xue2025opening,he2025viral,lin2025sim,luo2025emergent} or teleoperation \cite{ben2025homie,li2025clone,ze2025twist,ze2025twist2,li2025amo,fu2024humanplus,he2024omnih2o,ze2025generalizable,cheng2024open,cheng2024expressive,jiang2025wholebodyvla,luo2025sonic,liu2025trajbooster,wei2026psi_0}, both of which incur substantial overhead: sim-to-real requires intricate reward shaping and domain randomization \cite{peng2018sim,sadeghi2016cad2rl,tobin2017domain}, while teleoperation throughput is bottlenecked by operators, who must simultaneously stabilize the robot and compensate for tracking errors in real time. Recent teleoperation-free frameworks \cite{ding2025humanoid,yin2025visualmimic,yu2026bifrostumi,dong2026learning} mitigate these costs but remain confined to a narrow set of tasks in fixed environments. We instead present a portable, humanoid-free system for whole-body manipulation, verified on five tasks with generalization to unseen environments and objects.

\textbf{Teleoperation-free data collection.} This paradigm has proven highly efficient for fixed-base arms \cite{chi2024universal,rdt2,shafiullah2023bringing,song2020grasping,young2021visual,xu2025dexumi,choi2026wild,generalist2025gen0,hu2024data,lin2025onetwovla} and, more recently, for floating-base platforms such as quadrupeds \cite{ha2024umi} and aerial manipulators \cite{gupta2025umi}. However, these systems capture only end-effector trajectories, leaving humanoid whole-body coordination underspecified. \citet{xu2026hommi} extended the framework to mobile manipulation, but wheeled bases pose far fewer control challenges than legged humanoids. We present a teleoperation-free system tailored to humanoid whole-body manipulation.

\begin{figure}[!t]
    \centering
    \includegraphics[width=1\linewidth]{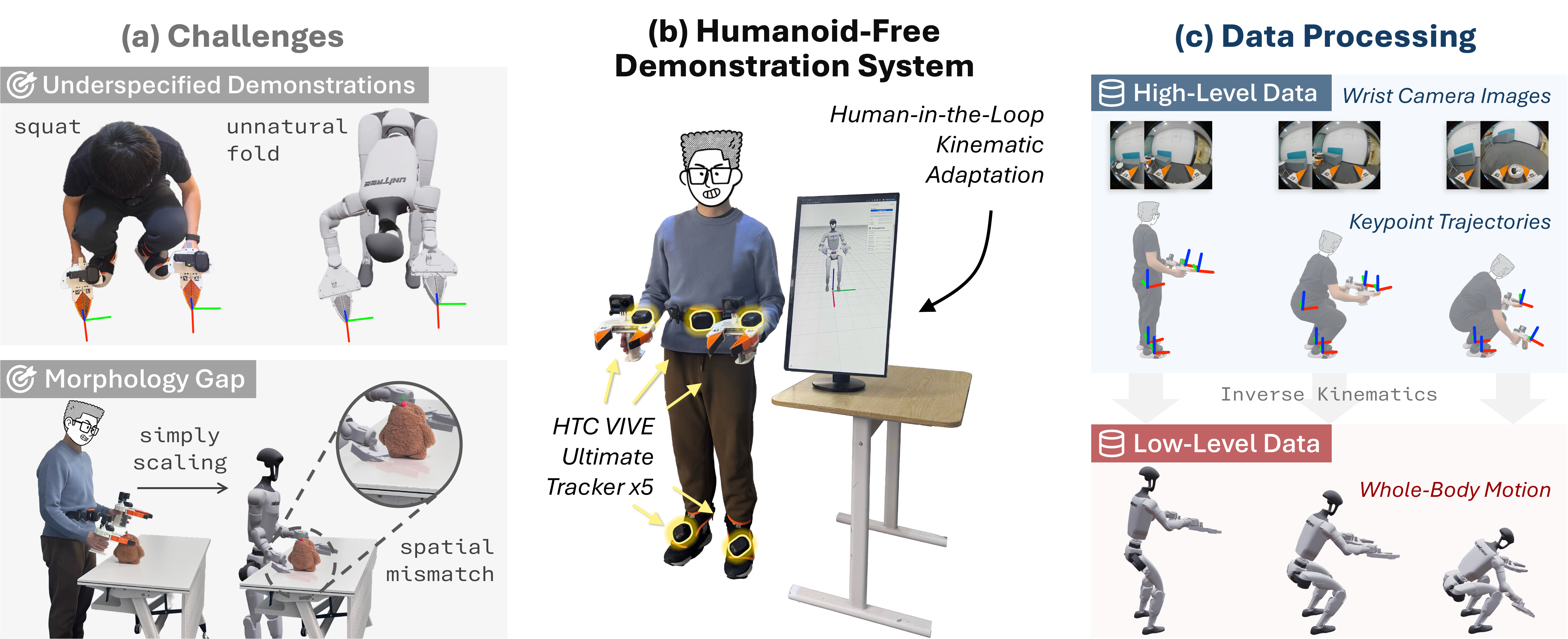}
    \caption{\textbf{HuMI data collection.} (a) \textbf{Challenges}: Gripper-only data under-specify whole-body motion, while naive scaling breaks scene alignment. (b) \textbf{Hardware}: Portable grippers and trackers with real-time IK preview to ensure feasibility. (c) \textbf{Data Processing}: Task-space $\mathrm{SE}(3)$ trajectories train the high-level policy, while whole-body IK solutions supervise the low-level controller.}
    \label{fig:data_collection}
\end{figure}

\section{Method}
\begin{figure}[!t]
    \centering
    \includegraphics[width=1\linewidth]{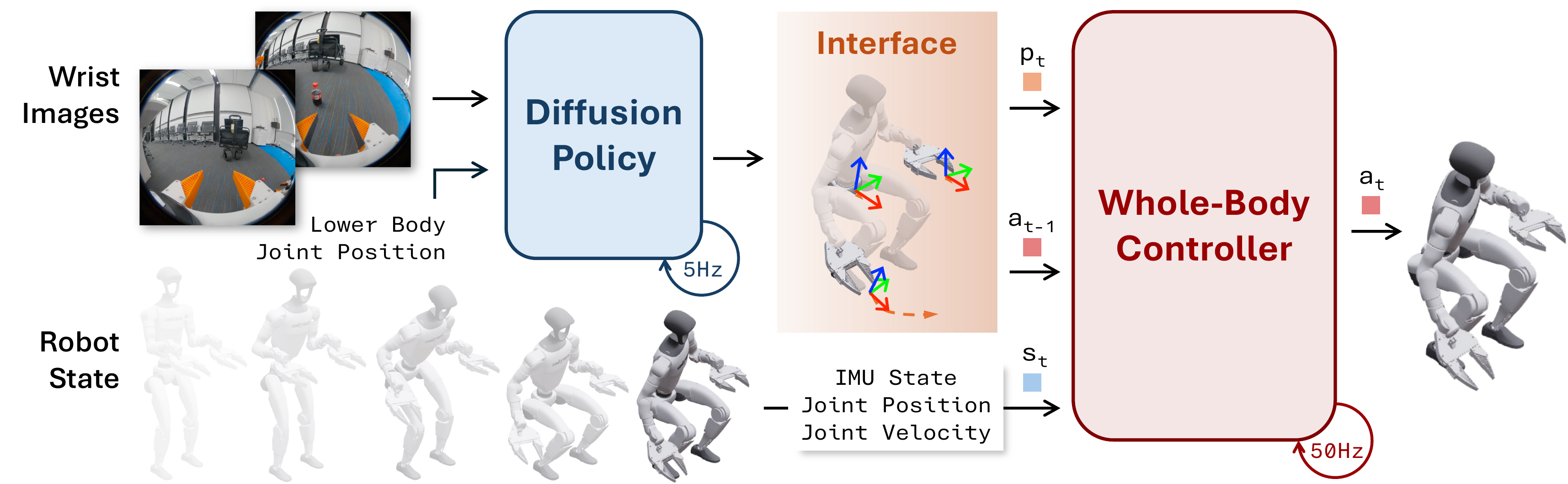}
    \caption{\textbf{HuMI control hierarchy.} A 5Hz Diffusion Policy maps images and proprioception to task-space keypoint chunks; a 50Hz whole-body controller converts these targets into joint actions.}
    \label{fig:policy_structure}
\end{figure}

HuMI consists of two components: a humanoid-free demonstration system that collects whole-body trajectories and image observations (Fig. \ref{fig:data_collection}), and a hierarchical policy learning framework (Fig. \ref{fig:policy_structure}). We use the collected demonstrations to train both levels: the high-level Diffusion Policy \cite{chi2025diffusion} maps observations to target keypoint trajectories, and the low-level controller learns joint actions for tracking such targets. At deployment, the two levels are integrated for closed-loop whole-body manipulation from onboard observations. We next describe each component and their integration.

\subsection{Humanoid-Free Demonstration System}
The primary goal of our demonstration system is to capture informative and robot-feasible human trajectories without requiring a physical humanoid. To this end, the system integrates portable, precise task-space recording hardware with a processing pipeline optimized for whole-body feasibility.

\textbf{Portable and precise hardware.}
The hardware design of HuMI prioritizes portability and precision while capturing sufficient whole-body information. We build upon UMI \cite{chi2024universal}, a widely adopted teleoperation-free system based on handheld grippers \cite{ha2024umi,gupta2025umi,xu2025dexumi,choi2026wild}. However, gripper trajectories alone under-specify whole-body motion, where torso and leg configurations can be task-critical. For instance, in low-reaching tasks, squatting and torso bending may produce similar end-effector poses, yet the latter can yield unnatural, collision-prone postures (Fig. \ref{fig:data_collection} (a) upper). HuMI therefore records five pose trajectories: two grippers, the pelvis, and two feet \cite{von2017sparse, chen2023full, heidicker2017influence}. To collect these trajectories portably, we use base-station-free HTC Vive Ultimate Trackers \cite{HTC2024VIVEUltimateTracker} on the corresponding frames, avoiding the overhead of Motion Capture systems \cite{Vicon2024Valkyrie,OptiTrack2024PrimeX41}. The resulting apparatus combines two 3D-printed handheld grippers with mounted GoPro cameras \cite{chi2024universal} and body trackers, yielding synchronized image observations and task-space $\mathrm{SE}(3)$ trajectories (Fig. \ref{fig:data_collection} (b)(c)).

\textbf{Human-in-the-loop kinematic adaptation.} 
A core challenge for HuMI is overcoming the embodiment gap between the human operator and the humanoid robot. Traditional retargeting methods often scale human motions to match the robot's morphology \cite{joao2025gmr,Luo2023PerpetualHC,yang2025omniretarget}. Yet for manipulation, such scaling changes the spatial relationship to fixed objects: scaling body height or arm length can cause insufficient reach or unintended intrusion (Fig. \ref{fig:data_collection} (a) bottom). Although interaction-mesh retargeting can preserve object-scene contacts \cite{yang2025omniretarget}, it breaks the visual-spatial alignment required by visuomotor policies. Therefore, HuMI tracks the original, unscaled poses from human trajectories.

Without scaling, however, these trajectories may become infeasible for the robot. Our target humanoid (Unitree G1 \cite{unitree2024g1}) is approximately 130\,cm tall, so adult-human motions can exceed its workspace or cause self-collisions near the body (see Appendix~\ref{sec:appendix_ik_interface}). We therefore provide a real-time IK preview of the virtual robot, letting operators adapt demonstrations to satisfy feasibility and task constraints (see Fig. \ref{fig:data_collection} (b)). Crucially, unlike physical teleoperation, which requires users to maintain balance and compensate for controller tracking errors \cite{ben2025homie,li2025clone,ze2025twist,ze2025twist2}, the preview only needs to ensure kinematic feasibility, substantially lowering cognitive load. Furthermore, the resulting IK solutions provide full-body joint supervision for low-level controller training (see Fig. \ref{fig:data_collection} (c)).

\subsection{Manipulation-Centric Whole-Body Controller}
To execute target trajectories from the high-level policy, we train an RL controller to track whole-body reference motions. Yet, state-of-the-art controllers \cite{luo2025sonic, liao2025beyondmimic} often incur tracking deviations of 4--6\,cm, which is insufficient for fine manipulation. A natural fix is to tighten end-effector (EE) tracking tolerance, but we find that this neglects whole-body coordination and compromises stability (see Appendix~\ref{sec:appendix_additional_experiments}). We address this with two mechanisms described below.

\textbf{Adaptive end-effector tracking.} 
To learn coarse coordinated motion, we first employ a basic whole-body tracking reward $r(\bar{e}_{\chi}, \sigma_{\chi}) = \exp(-\bar{e}_{\chi} / \sigma_{\chi}^2)$ following standard practice. For each metric $\chi \in \{\mathbf{p}, \theta, v, w\}$ (position, orientation, and linear/angular velocity), $\bar{e}_{\chi}$ denotes the mean squared tracking error \cite{liao2025beyondmimic} and $\sigma_{\chi}$ its tolerance. The total whole-body tracking reward is then:
\begin{equation*}
r_{\text{body}} = \sum_{\chi \in \{\mathbf{p}, \theta, v, w\}} r(\bar{e}_{\chi}^{\text{body}}, \sigma_{\chi}).
\end{equation*}

Beyond whole-body coordination, manipulation further requires phase-dependent end-effector precision. Consider a human kneeling to pick up an object: the initial rapid descent tolerates looser tracking, whereas the slower final grasp requires precise contact. Inspired by this intuition, we therefore scale end-effector tolerance with speed, enforcing high accuracy during slow interactions and allowing greater flexibility as velocity increases. The adaptive end-effector reward is defined as:
\begin{equation*}
r_{\text{EE}} = \mathbb{I}\left(\left\|v^\text{ref}_{\text{base}}\right\| < \delta\right) \cdot \sum_{\chi \in \{\mathbf{p}, \theta\}} r\left(\bar{e}_{\chi}^{\text{EE}}, \sigma_{\chi}\left(v^\text{ref}_{\text{EE}}\right)\right),
\end{equation*}
where $\sigma_{\chi}(v^\text{ref}_{\text{EE}})$ interpolates between $[\sigma_{\chi}^{\min}, \sigma_{\chi}^{\max}]$ according to reference EE speed, with a base-velocity gate for stability during rapid movement. Combining the whole-body and end-effector objectives gives $r_{\text{tracking}} = w_{\text{body}}r_{\text{body}} + w_{\text{EE}}r_{\text{EE}}$. We also find that a curriculum on $w_{\text{EE}}$ and $\sigma_{\chi}^{\min}$ is necessary to delay EE precision emphasis until stable global motion is learned (see Appendix~\ref{sec:appendix_training_details}).

\textbf{Variable-speed augmentation.}
In standard motion tracking, the reference typically advances at a fixed speed. In this case, the target often moves on too fast before the policy can spend enough time fixing small mistakes, making it hard to learn highly precise movements. We therefore introduce a variable execution pace to overcome this limitation. For a reference motion with duration $T$, we scale the execution speed within $[s_{\min}, s_{\max}]$ by sampling a new speed scaling factor $s_k$ every $\Delta$ seconds (see Appendix~\ref{sec:appendix_training_details} for details). This variety of slow speeds within each episode gives the policy ample time to fix small errors, thereby facilitating the learning of high-precision movements.

\subsection{Policy Interface for Improved System Integration}
\begin{wrapfigure}{r}{0.42\linewidth}
    \centering
    \includegraphics[width=\linewidth]{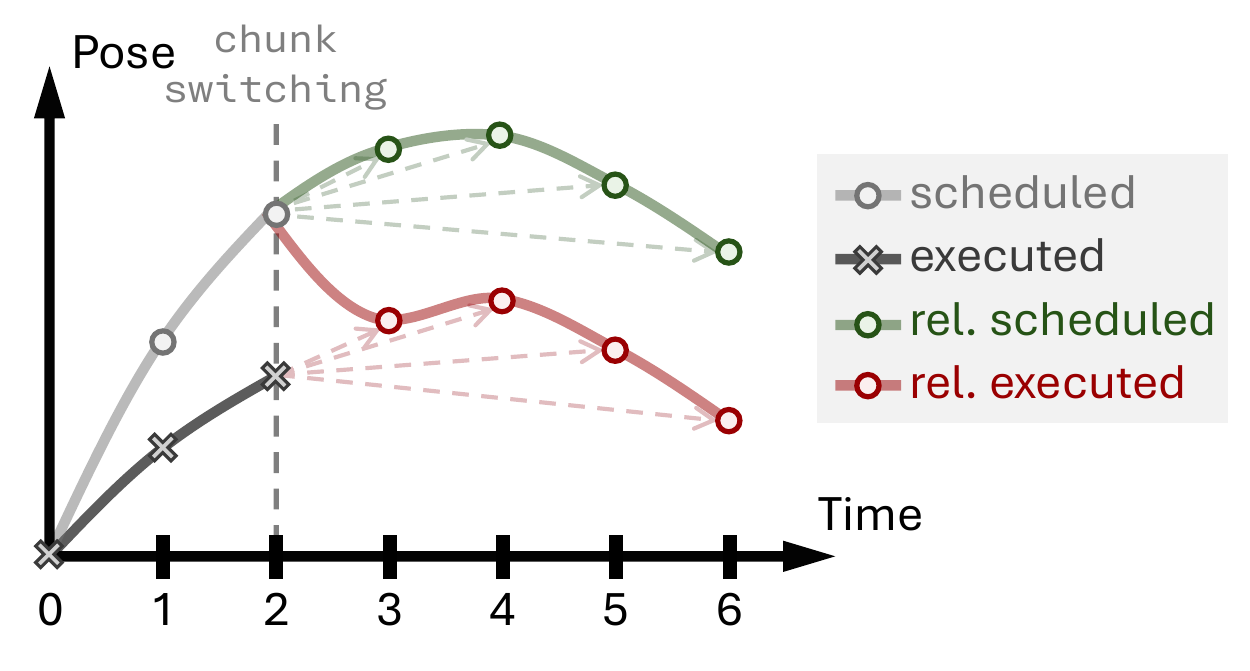}
    \caption{\textbf{Impact of reference frame selection on tracking-induced chunk boundary offsets.}}
    \label{fig:relative_chunk}
\end{wrapfigure}
As shown in Fig.~\ref{fig:policy_structure}, the high-level Diffusion Policy \cite{chi2025diffusion} predicts action chunks represented as relative keypoint trajectories \cite{chi2024universal,ha2024umi,gupta2025umi}. Training details are provided in Appendix~\ref{sec:appendix_highlevel_training_details}. However, naively feeding these targets to the low-level controller is fragile: coupled errors from both levels can significantly compromise stability. We therefore introduce two policy-interface modifications for robust whole-body execution.

\noindent\textbf{Target pose as high-level action reference.}
Prior manipulation policies typically use the \textit{executed} EE pose as reference for new chunks \cite{chi2024universal,ha2024umi,gupta2025umi}. For humanoid manipulation, however, low-level tracking errors remain non-negligible even with improved tracking, causing the \textit{executed} pose to lag behind the \textit{scheduled} target at chunk boundaries. As shown in Fig.~\ref{fig:relative_chunk}, resetting the next chunk to this lagging pose induces reversals at chunk boundaries (\textcolor{humiFigRed}{red line}). HuMI instead uses the previous \textit{scheduled} pose as the action reference, avoiding tracking-induced boundary offsets (\textcolor{humiFigGreen}{green line}). This also matches training: the high-level policy assumes perfect tracking (as it is trained via imitation learning on human trajectories), while the low-level RL controller is trained to track fixed offline references.

\begin{wrapfigure}{r}{0.58\linewidth}
    \centering
    \includegraphics[width=\linewidth]{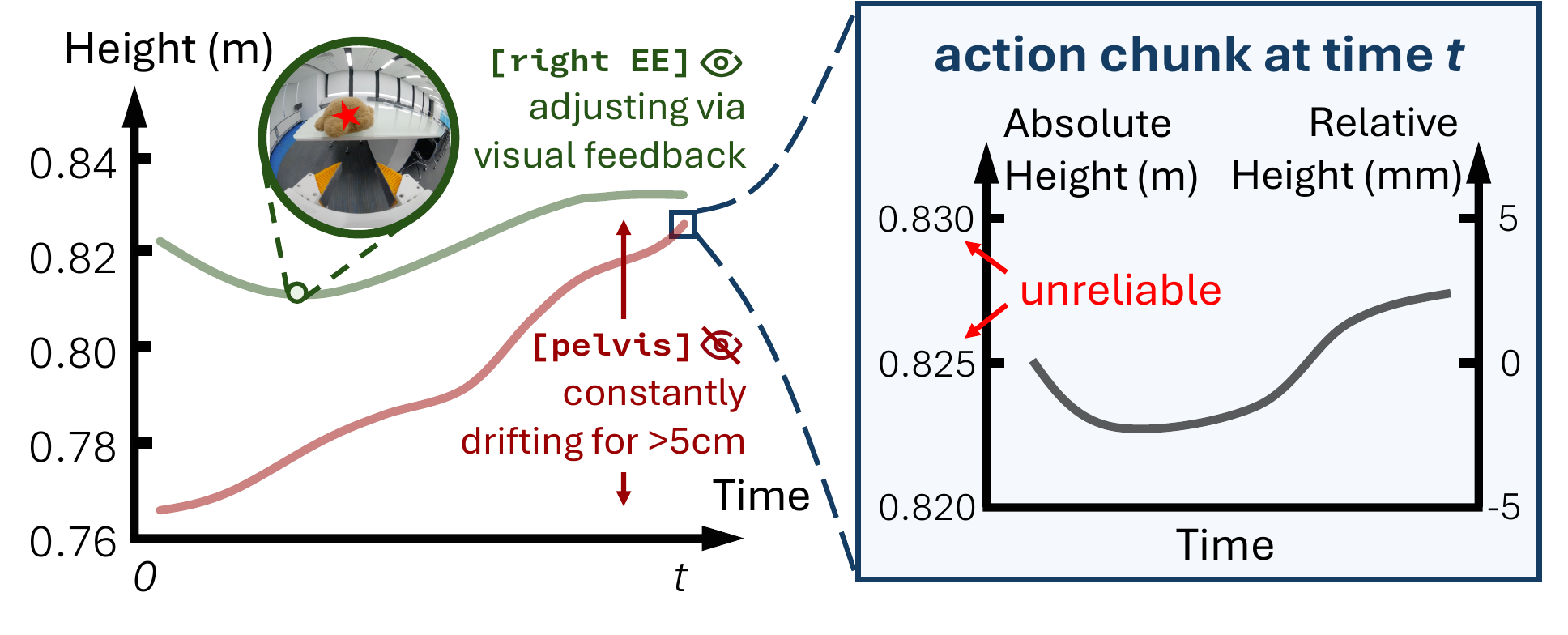}
    \caption{\textbf{Within-chunk relative tracking mitigates drift in non-vision-grounded keypoints.}}
    \label{fig:relative_blind}
\end{wrapfigure}
\noindent\textbf{Relative pose tracking for non-vision-grounded keypoints.}
Full-body keypoints provide richer motion information, but controlling more keypoints introduces an observability trade-off. Fig.~\ref{fig:relative_blind} shows a stationary grasping task: grippers receive wrist-view visual feedback and can be adjusted in subsequent chunks, whereas keypoints such as the pelvis and feet are ``blind'', so drift in their absolute targets can accumulate significantly ($>$5\,cm). We therefore track the within-chunk relative transform for non-vision-grounded keypoints rather than the absolute transform (Fig.~\ref{fig:relative_blind} right). This enables flexible keypoint configurations; we use a 3-keypoint setup (grippers + pelvis) by default and demonstrate that the system maintains robust performance even when scaled to 5 keypoints (adding feet).

\section{Experiments}
In this section, we empirically evaluate HuMI along three key dimensions. Specifically, we aim to answer the following questions:
\begin{enumerate}
    \item \textbf{Whole-body manipulation capability.} Can HuMI learn feasible whole-body skills from humanoid-free demonstrations with accurate end-effector tracking, while effectively coordinating high- and low-level policies during autonomous execution?
    \item \textbf{Generalization ability.} Do humanoid-free demonstrations collected across varied environments enable the learned policy to generalize to unseen environments and objects?
    \item \textbf{Data-collection efficiency.} Can HuMI acquire whole-body manipulation data efficiently and with a high acceptance rate?
\end{enumerate}

To answer these questions, we evaluate HuMI on five representative whole-body manipulation tasks under in-domain settings and out-of-domain environment/object shifts. We further compare its data-collection efficiency against TWIST2~\cite{ze2025twist2}, a state-of-the-art humanoid teleoperation system.

\begin{figure}[!t]
    \centering
    \includegraphics[width=1\linewidth]{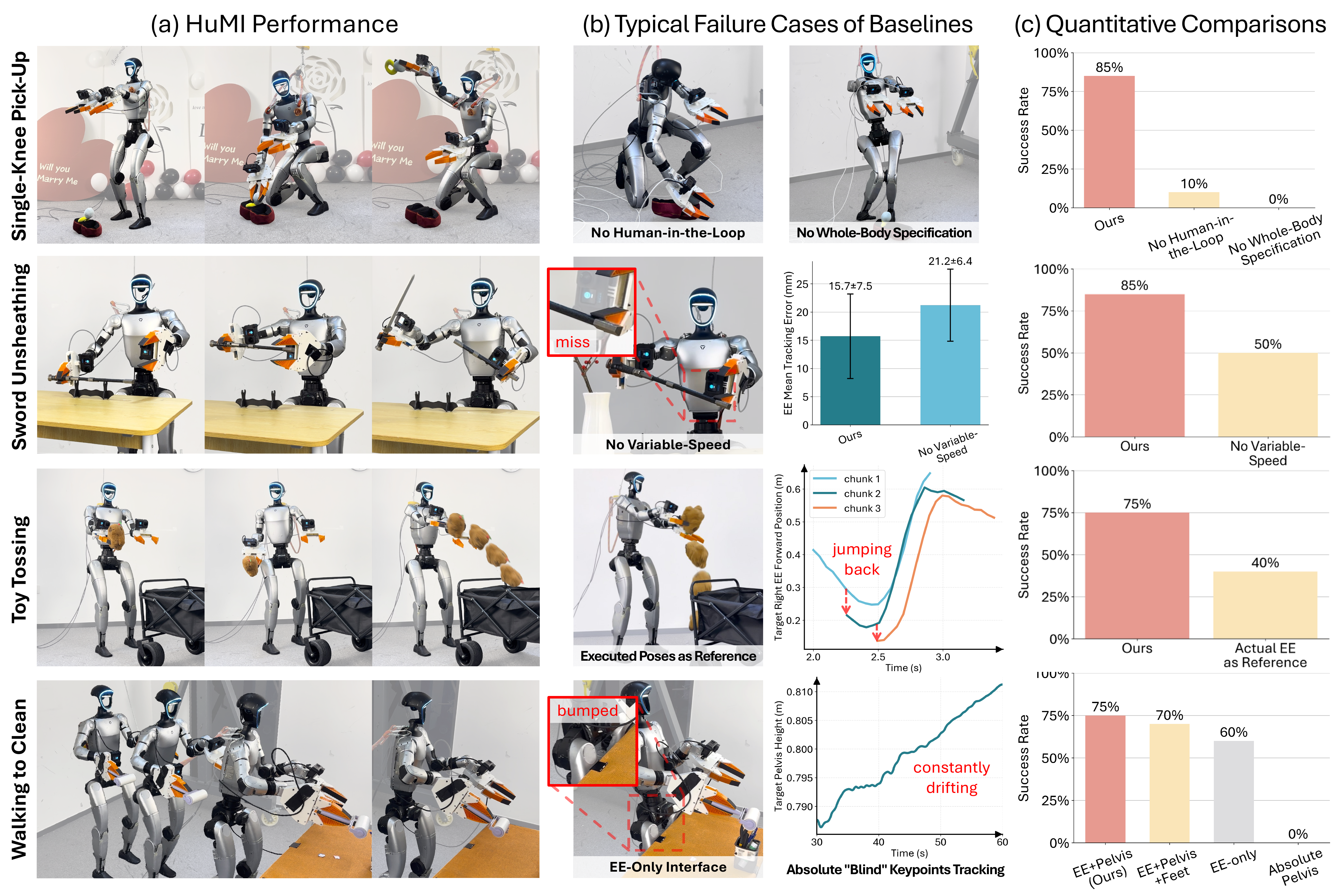}
    \caption{\textbf{Whole-body capability results.} (a) \textbf{HuMI performance} on four tasks. (b) \textbf{Baseline failure modes}; red boxes highlight the failure behavior details. (c) \textbf{Success rates} for each task.}
    \label{fig:experiment_results}
\end{figure}
\subsection{Whole-Body Manipulation Capability}
The capability study covers four tasks spanning complementary capabilities: whole-body coordination, bimanual control, high-speed dynamics, and loco-manipulation. All experiments use in-domain settings, with the same environments and initial robot--object configurations as data collection.
\subsubsection{Learning Feasible Whole-Body Skills}
We evaluate HuMI on a \textbf{single-knee pick-up} task: the robot kneels from standing onto its right knee, grasps a ring toy on the ground with its right hand, and lifts it to a target pose.

\underline{\textbf{Task challenges.}} The motion requires coordinated posture control to maintain balance over a narrow support region while reaching near the ground. The final grasp further requires accurate end-effector placement despite large whole-body displacement.\\
\underline{\textbf{Performance.}} HuMI succeeds in $17/20=85\%$ trials (Fig.~\ref{fig:experiment_results} (c)), producing stable kneeling and lifting motions in successful rollouts (Fig.~\ref{fig:experiment_results} (a)).\\
\textbf{\underline{Remove human-in-the-loop kinematic adaptation.}} We train an ablated policy on demonstrations collected without the online humanoid-avatar preview. Success drops to $1/10=10\%$, and rollouts often contain kinematically inappropriate kneeling, such as an excessively splayed leg (Fig.~\ref{fig:experiment_results} (b)). This suggests that kinematic feedback helps operators keep demonstrations within the target humanoid's feasible motion space.\\
\textbf{\underline{Remove whole-body specification.}} We also test an EE-only interface following UMI-style systems~\cite{chi2024universal,ha2024umi,gupta2025umi}, where the high-level policy commands only end-effector waypoints and the low-level controller is trained from these sparse waypoints. Success drops to $0/10$, indicating that EE-only supervision under-specifies the coordinated whole-body motion required for this task.
\subsubsection{Humanoid Bimanual Manipulation}
To evaluate HuMI’s ability to coordinate two hands during a manipulation task, we consider a \textbf{sword-unsheathing} task. In this task, the robot grasps the hilt on the rack, supports the scabbard with its left hand, and coordinates both arms to draw the blade.

\underline{\textbf{Task challenges.}} This task requires stable contacts, appropriate end-effector placement, and coordinated motion of both arms. The sword hilt and scabbard provide limited grasping areas, so pose errors can destabilize contact or lead to collisions with the rack. Once both hands are engaged, mismatches in timing or motion can introduce shear that degrades alignment, making this task a useful test case for humanoid bimanual coordination.\\
\underline{\textbf{Performance.}} HuMI completes the unsheathing task in $17/20=85\%$ trials, with an average end-effector tracking error of 15.7\,mm (Fig.~\ref{fig:experiment_results} (c)). In successful trials, the robot establishes stable contacts with the hilt and scabbard while avoiding visible slippage or rack collisions. It then coordinates both arms sufficiently to draw the blade in a continuous motion (Fig.~\ref{fig:experiment_results} (a)).\\
\underline{\textbf{Remove variable-speed augmentation.}} We ablate the variable-speed augmentation by training the low-level policy on demonstrations replayed only at original human speeds. The success rate drops from $85\%$ to $5/10=50\%$, with the average tracking error increasing to 21.2\,mm (Fig.~\ref{fig:experiment_results} (b)). The policy often fails to establish a stable grasp or shows less consistent bimanual coordination. These degradations suggest that variable-speed augmentation helps the controller reduce tracking errors and coordinate the two arms during the task.
\subsubsection{Temporally Coherent Dynamic Control}
The third experiment uses a \textbf{toy-tossing} task to evaluate whether HuMI can transfer high-speed human motions to a humanoid. The robot must produce a temporally structured throw, with a backward wind-up followed by a rapid forward swing toward a target container.

\underline{\textbf{Task challenges.}} The task requires coordinating the torso and arms to route momentum through coupled joints. It also stresses the hierarchy: the high-level policy must issue temporally coherent trajectory chunks, and the low-level policy must track them as whole-body motion.\\
\underline{\textbf{Performance.}} HuMI completes the task in $15/20=75\%$ trials (Fig.~\ref{fig:experiment_results} (c)). In successful trials, the robot follows smooth whole-body trajectories and releases the object near the forward-swing peak, yielding sufficient speed and direction for the object to land in the container (Fig.~\ref{fig:experiment_results} (a)).\\
\underline{\textbf{Executed EE poses as action reference.}} We ablate the action reference by replacing the previous scheduled EE pose with the executed EE pose. This reduces success to $4/10=40\%$ (Fig.~\ref{fig:experiment_results} (b)(c)). In failed trials, the end-effector trajectory becomes less continuous across chunks, and reversals reduce the forward-swing momentum, often leading to lower release speed or inaccurate direction.
\subsubsection{Loco-Manipulation}
We evaluate loco-manipulation with a \textbf{walking-to-clean} task. Each rollout starts with the robot 1--2\,m from the target desk and an initial yaw offset sampled from $[-45^\circ, 45^\circ]$. The robot must approach the desk and use a lint roller to clear scattered paper scraps from the tabletop.

\underline{\textbf{Task challenges.}} This task couples locomotion with fine tabletop manipulation. The policy must first guide stable footsteps toward the desk, then transition to wiping once the surface is reachable. This transition requires the interface to convey both locomotion-oriented and manipulation-oriented intent, so that the final stance leaves the tabletop within the arm's workspace.\\
\underline{\textbf{Performance.}} HuMI completes $15/20=75\%$ rollouts (Fig.~\ref{fig:experiment_results} (c)), approaching from varied initial poses and clearing the tabletop scraps with overlapping lint-roller strokes (Fig.~\ref{fig:experiment_results} (a)).\\
\textbf{\underline{Design of keypoints interface.}}
We ablate the keypoint set passed from the high-level policy to the low-level controller. The compared interfaces differ only in which task-space trajectories are commanded: \textbf{EE-only} uses the two end-effector (gripper) poses; \textbf{EE+pelvis} additionally includes the pelvis/floating-base pose; and \textbf{EE+pelvis+feet} further includes the two foot poses. The EE-only interface reaches $6/10=60\%$ success (Fig.~\ref{fig:experiment_results} (b)(c)), with failures mainly during approach: the robot stops too far from the desk or collides with it. These failures suggest that end-effector targets alone can be ambiguous between stepping forward and reaching forward. In contrast, the two interfaces with additional body keypoints show comparable behavior in our trials: EE+pelvis achieves $15/20=75\%$ success, and EE+pelvis+feet achieves $7/10=70\%$. We therefore treat both as viable interfaces within HuMI.\\
\uline{\textbf{Absolute pose tracking for non-vision-grounded keypoints.}}
We also replace relative pelvis tracking with absolute tracking in the global frame under the EE+pelvis interface. Success drops from $15/20=75\%$ to $0/10$, as accumulated drift in pelvis targets cannot be corrected without visual feedback, causing the robot to veer off course and lose balance (Fig.~\ref{fig:experiment_results} (b)).
\subsection{Generalization Ability}
\label{sec:exp_generalization}
We evaluate generalization on a \textbf{squat-and-retrieve} task: the humanoid localizes a floor-placed bottle, squats, grasps it, and lifts it. We test two distribution shifts: unseen environments and objects.

\textbf{\uline{Training data.}} We collect 350 whole-body demonstrations across 7 environments and 7 bottle instances (Appendix~\ref{sec:appendix_generalization_experiments}). These demonstrations cover variations in scene layout, lighting, floor appearance, and object appearance, and are used to train a diffusion policy for humanoid control.\\
\uline{\textbf{Generalization settings.}} We evaluate the policy with both environments and objects held out in each rollout (see Appendix~\ref{sec:appendix_generalization_experiments}). For \textbf{environment generalization}, the humanoid is deployed in four environments not used during training, including scenes with different layouts, clutter, and illumination. This setting tests whether the policy can use visual observations to identify task-relevant cues despite background changes. For \textbf{object generalization}, we test on six held-out objects, including bottles and bottle-like objects absent from the training set, to evaluate robustness to changes in object shape, size, and texture. Across 20 trials, HuMI completes $14/20=70\%$ episodes, suggesting that diverse humanoid-free demonstrations can support environment- and object-level generalization on this representative task.
\subsection{Data Collection Efficiency}
\begin{wrapfigure}{r}{0.55\linewidth}
    \centering
    \includegraphics[width=\linewidth]{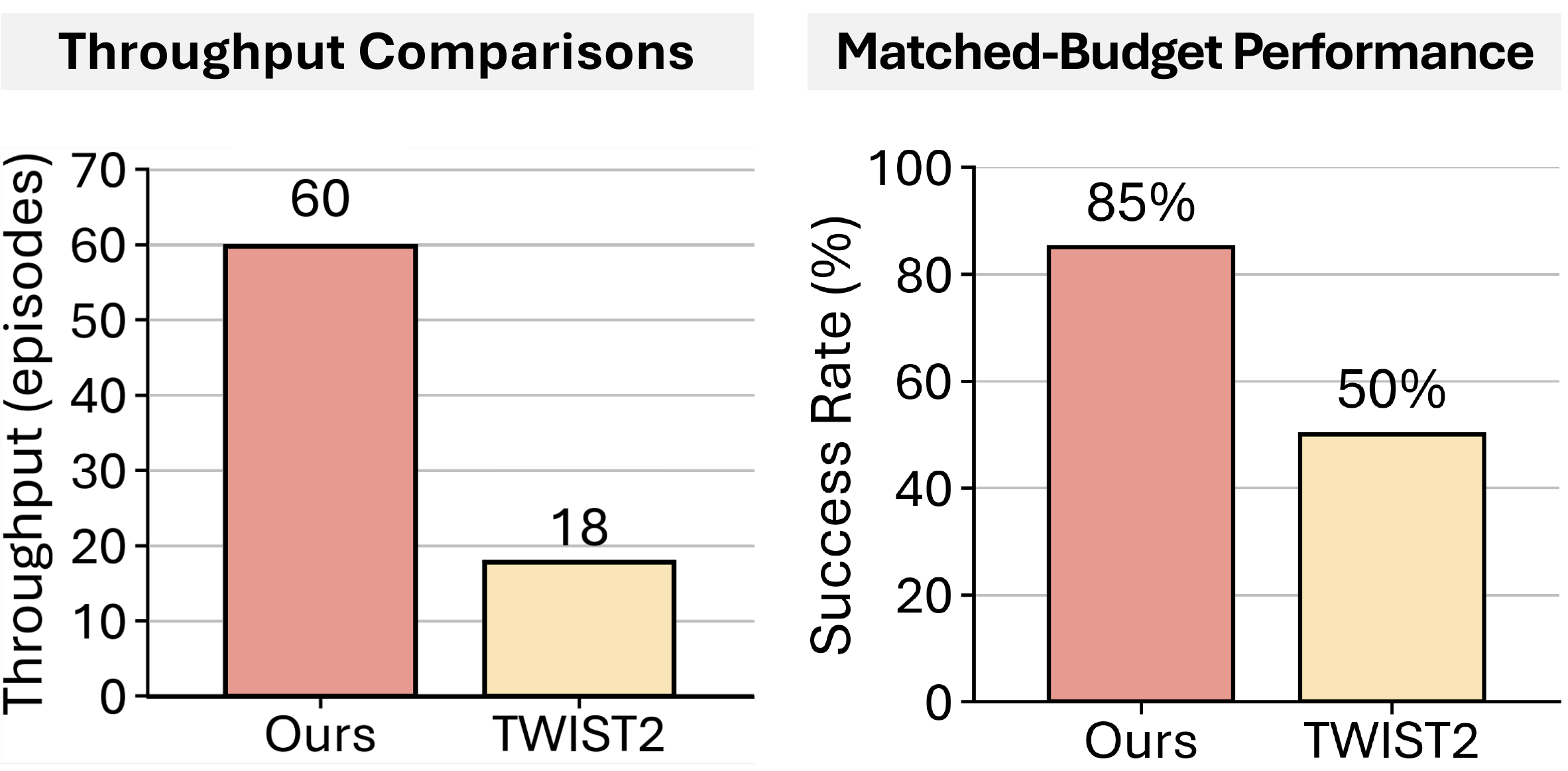}
    \caption{\textbf{Data collection efficiency comparison.}}
    \label{fig:twist_comparison}
\end{wrapfigure}
We compare HuMI with the TWIST2 teleoperation workflow~\cite{ze2025twist2} on the sword-unsheathing task using 15\,min collection sessions. Experienced users (with $>$10\,h of experience on each system) operate both systems, and we measure the number of valid training demonstrations collected in each session. As shown in Fig.~\ref{fig:twist_comparison}, HuMI collects \textbf{60} valid demonstrations compared with \textbf{18} for TWIST2, yielding over 3$\times$ the valid-demo throughput of TWIST2.

To enable a comparison under a matched data budget, we also collect TWIST2 demonstrations for the same total demonstration time as HuMI in the main experiment ($\sim$22\,min). Under this setting, HuMI achieves an 85\% success rate, compared with 50\% for TWIST2 teleoperation. This result suggests that HuMI compares favorably with teleoperation given the same total demonstration time. For a broader and fairer comparison, we additionally consider the bimanual pick-and-place task from the TWIST2 paper and a single-knee pick-up task requiring more whole-body coordination, with detailed protocols and results reserved for Appendix~\ref{sec:appendix_additional_experiments}.

\section{Conclusions and Limitations}
We presented HuMI, a framework that bridges portable humanoid-free demonstration collection and humanoid whole-body execution through a coordinated policy-controller hierarchy. Experiments show a 3$\times$ data collection gain over teleoperation and 70\% success in unseen environments.

\textbf{Limitations.} HuMI has several limitations. (1) Our motion-capture pipeline assumes sufficient environmental texture and lighting for its base-station-free visual trackers \cite{HTC2024VIVEUltimateTracker}; in low-texture or poorly-lit scenes, tracker dropout or drift can corrupt demonstrations. As such tracking systems continue to improve, HuMI could become applicable to a broader range of everyday environments. (2) The low-level controllers can occasionally lose balance and are trained per task rather than as a single general-purpose policy. Future work could train a unified general-purpose low-level controller that subsumes the current per-task variants. (3) Our generalization study covers unseen environments and objects but not unseen embodiments, and evaluation is confined to a single humanoid platform \cite{unitree2024g1}; we plan to extend evaluation to additional humanoid platforms.

\clearpage
\acknowledgments{%
This research was conducted with the support of the Shanghai Qi Zhi Institute \& Spirit AI Innovation Program. Funding and support for this work were also provided by the Tsinghua University - Keystone Electrical (Zhejiang) Co., Ltd. Joint Research Center for Embodied Multimodal Artificial Intelligence (JCEMAI). Additionally, we would like to extend our thanks to the Xiongan AI Institute. This project is also supported by Beijing Natural Science Foundation (JR25019).

We thank Fanqi Lin and Gu Zhang for their contributions to the conceptual development. We also thank Chengbo Yuan and Rui Zhou for early discussions regarding the hardware setup, and Haoyang Weng for assistance with controller training. Finally, we are grateful to Jiacheng You for invaluable discussions concerning the experimental design.
}

\bibliography{references}

\clearpage
\appendix
\raggedbottom

\section*{Appendix}
Please visit the project website \url{https://humanoid-manipulation-interface.github.io/} to view the real-world robot rollout videos.

\section*{Appendix Contents}
\newcommand{\appendixcontentsline}[2]{%
    \noindent\hyperref[#1]{\ref*{#1}\quad #2}\dotfill\hyperref[#1]{\pageref*{#1}}\par
}
\begingroup
\setlength{\parskip}{2pt}
\appendixcontentsline{sec:appendix_data_collection_system}{Data Collection System Details}
\appendixcontentsline{sec:appendix_deployment_details}{Deployment Details}
\appendixcontentsline{sec:appendix_additional_experiments}{Additional Experiments}
\appendixcontentsline{sec:appendix_generalization_experiments}{Generalization Experiments Details}
\appendixcontentsline{sec:appendix_training_details}{Low-Level Controller Training Details}
\appendixcontentsline{sec:appendix_highlevel_training_details}{High-Level Policy Training Details}
\endgroup

\section{Data Collection System Details}
\label{sec:appendix_data_collection_system}
This section provides details of our data collection system. We first describe the hardware components, followed by the human-in-the-loop IK adaptation interface, the data processing pipeline, and finally, the data collection protocol.

\subsection{Hardware Components}
\label{sec:appendix_hardware_components}
\begin{figure}[h]
    \centering
    \includegraphics[width=.65\linewidth]{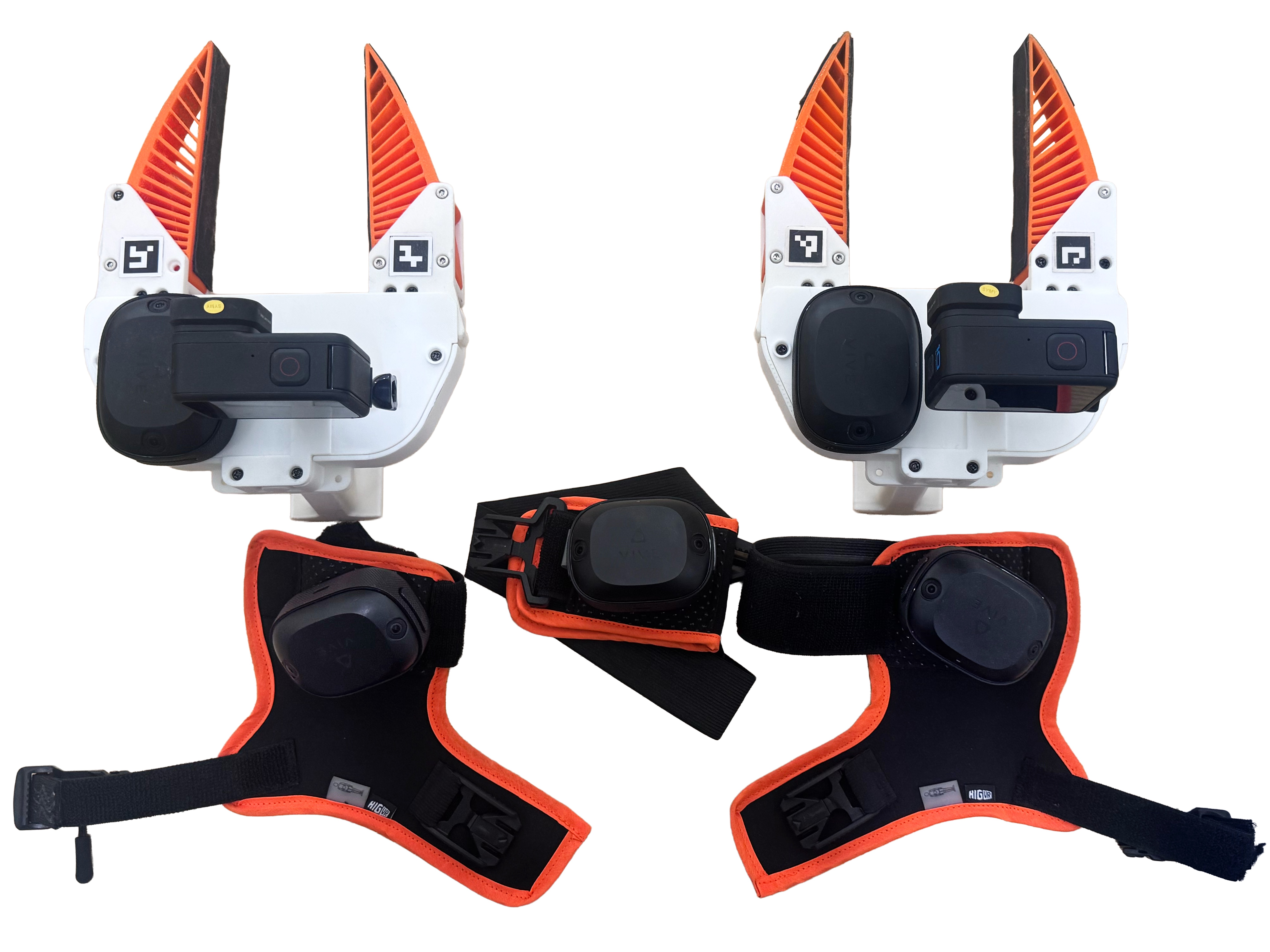}
    \caption{\textbf{HuMI's hardware setup}.}
    \label{fig:appendix_data_collection_hardware}
\end{figure}
As illustrated in Fig. \ref{fig:appendix_data_collection_hardware}, our data collection system comprises the following components:
\begin{itemize}
    \item \textbf{Two UMI \cite{chi2024universal} grippers}: We utilize UMI grippers equipped with GoPro cameras \cite{GoProHero10} to record wrist-view RGB observations and ArUco markers \cite{romero2018speeded,garrido2016generation} for gripper width tracking.
    \item \textbf{Five HTC VIVE Ultimate trackers \cite{HTC2024VIVEUltimateTracker}}: To capture 6-DoF poses, we attach trackers to the two grippers, the waist, and the feet. We modified the top cover of the original UMI gripper design to accommodate a tracker mount. Standard VIVE straps are used to secure the trackers to the waist and feet.
\end{itemize}
A laptop running SteamVR \cite{SteamVR} is required to interface with the trackers to record their poses.

\subsection{IK Interface}
\label{sec:appendix_ik_interface}
\begin{figure}[H]
    \centering
    \includegraphics[width=\linewidth]{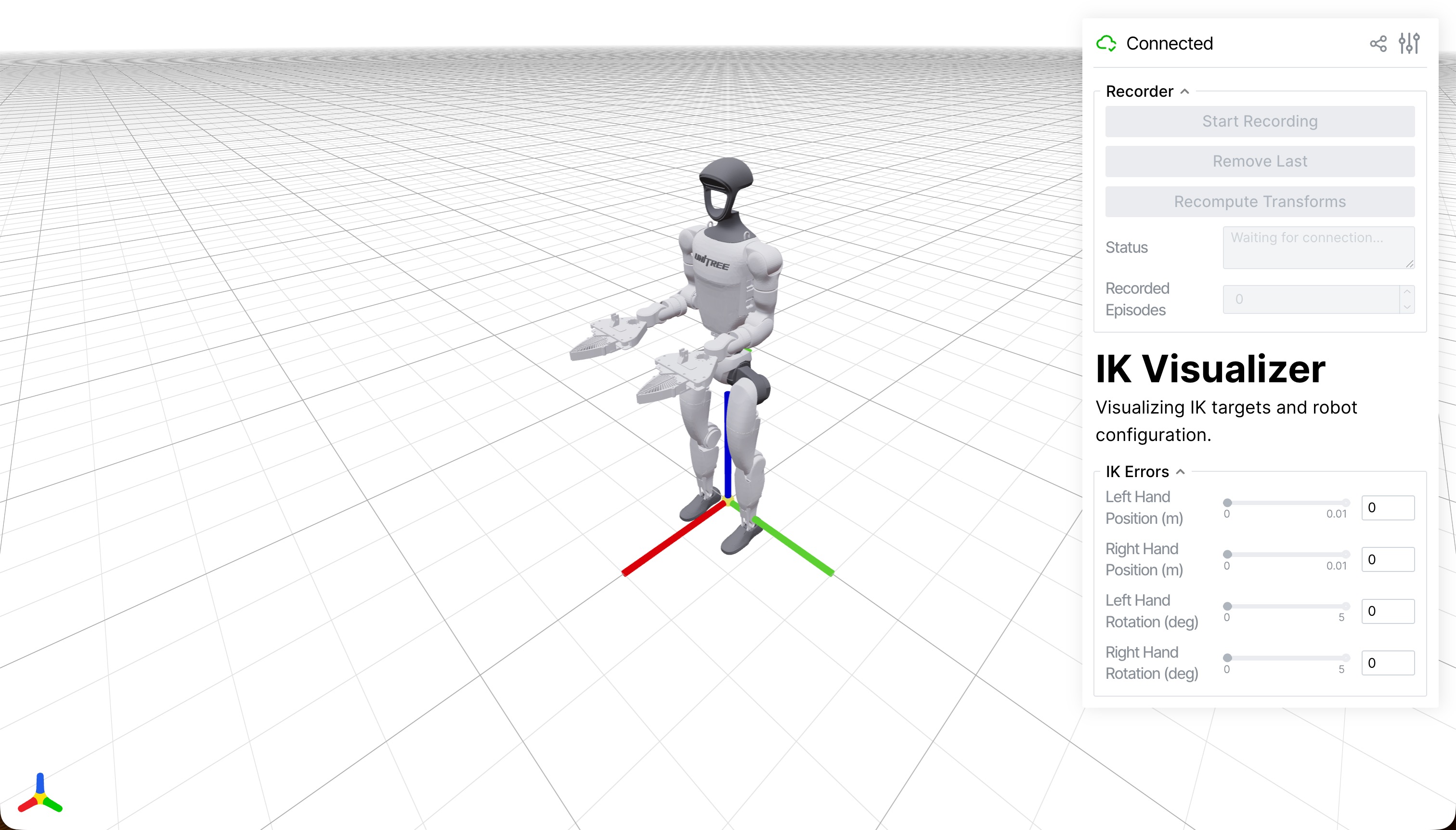}
    \caption{\textbf{IK preview interface}.}
    \label{fig:appendix_ik_interface}
\end{figure}

Unscaled human trajectories preserve task geometry, but they can be kinematically infeasible for the target humanoid when the human workspace exceeds the robot workspace or induces self-collisions. To address this, we developed a real-time IK preview interface that visualizes how the target humanoid tracks the current human keypoints during collection. The demonstrator can then adjust their posture before accepting the trajectory, rather than relying on post-hoc motion scaling. Notably, to preserve strict spatial relationships between the robot and the environment, we use the original, unscaled motions (specifically, the trackers' recorded $\mathrm{SE}(3)$ transforms in the world frame) as IK targets. Scaling is applied exclusively to the height of the pelvis tracker.

The IK problem incorporates three subtasks: tracking the translation and rotation of the five keypoints, avoiding self-collisions, and maintaining a natural configuration via posture regularization. We solve the IK problem in real-time using Mink \cite{Zakka_Mink_Python_inverse_2025} and visualize the robot's configuration using Viser \cite{yi2025viser}. Fig. \ref{fig:appendix_ik_interface} depicts the user interface during data collection.

\vspace{0.4em}
\begin{figure}[H]
    \centering
    \captionsetup{skip=2pt}
    \captionsetup[subfigure]{skip=1pt}
    \newcommand{\ikpreviewfigureheight}{0.16\textheight}
    \begin{subfigure}[t]{0.42\linewidth}
        \centering
        \includegraphics[height=\ikpreviewfigureheight]{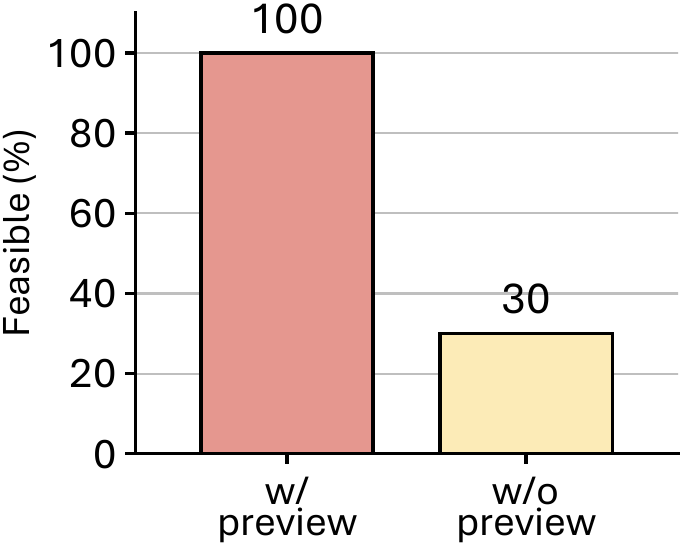}
        \caption{Feasible trajectory rate.}
    \end{subfigure}
    \hfill
    \begin{subfigure}[t]{0.54\linewidth}
        \centering
        \includegraphics[height=\ikpreviewfigureheight]{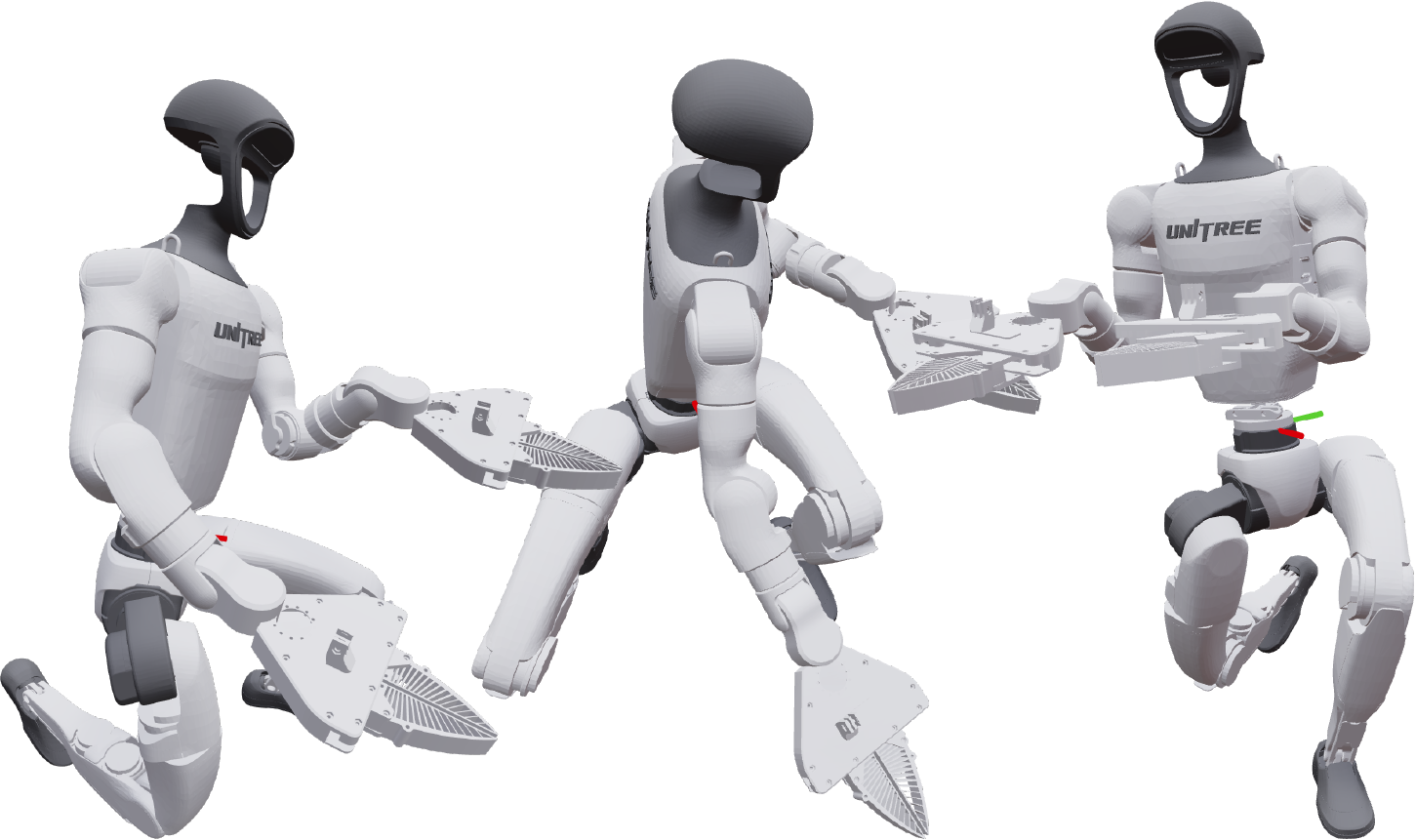}
        \caption{Infeasible motion frames without preview.}
    \end{subfigure}
    \caption{\textbf{Effect of online IK preview}. Without online preview, unscaled human motions frequently exceed the target humanoid's kinematic workspace or cause self-collisions.}
    \label{fig:appendix_infeasible_motions}
\end{figure}

We quantify this adaptation on the \textbf{single-knee pick-up} task using 20 demonstrations collected without IK preview. Only \textbf{30\%} of these demonstrations were kinematically feasible; the remaining \textbf{70\% (14/20)} contained infeasible frames (Fig.~\ref{fig:appendix_infeasible_motions}). With online preview, demonstrators adapted their posture: the average right hand-to-pelvis distance increased from \textbf{76.7 cm} to \textbf{90.7 cm}, moving the grasping hand away from compact, self-colliding poses and eliminating the observed kinematic failures. Since the IK targets remain unscaled, task geometry is preserved up to negligible tracking error (\textbf{1.7 mm} / $\mathbf{6.6\times10^{-4}}$ degrees).

\subsection{Data Processing}
\label{sec:appendix_data_processing}
The recorded data consists of MP4 videos from the gripper cameras and time-stamped $\mathrm{SE}(3)$ trajectories from the five trackers. We process the data in the following steps:
\begin{enumerate}
    \item \textbf{Synchronization}: We use the tracker timestamps as the reference clock since the five trackers are natively synchronized. To align the gripper videos with the tracker data, we extract the gyroscope data embedded in the video files by the GoPro cameras. We then align the video and tracker timestamps by cross-correlating the magnitudes of their angular velocities.
    \item \textbf{Gripper width extraction}: Following \citet{chi2024universal}, we extract the gripper width from the recorded videos by detecting the ArUco markers attached to the grippers. 
    \item \textbf{Data packaging}: The recorded data are packaged into two subsets: (1) visual observations, gripper widths, and keypoint trajectories for training the high-level policy; and (2) keypoint trajectories paired with whole-body IK solutions for training the low-level controller.
\end{enumerate}

\subsection{Data Collection Protocol}
\label{sec:appendix_data_collection_protocol}
The following is the step-by-step protocol for collecting demonstrations in a new scene:
\begin{enumerate}
    \item \textbf{Mapping setup}: For each new scene, the demonstrator must follow VIVE's prompts to build a tracking map of the environment; this typically takes 1--2 minutes.
    \item \textbf{Calibration and synchronization}: The demonstrator may optionally perform gripper width calibration and GoPro timestamp synchronization following \citet{chi2024universal}.
    \item \textbf{Demonstration}: The demonstrator repeatedly performs the task within the scene. We use the start and stop times of the recorded videos to delineate episodes. The demonstrator uses GoPro's voice commands to control video recording, while the tracker data recording is controlled via the IK interface GUI (Fig. \ref{fig:appendix_ik_interface}). Thanks to the synchronization step described in Sec. \ref{sec:appendix_data_processing}, strict alignment of start/stop times between the videos and tracker recordings is not required. During the demonstration, the demonstrator adapts their motions based on the real-time IK preview. 
\end{enumerate}

\section{Deployment Details}
\label{sec:appendix_deployment_details}

In this section, we detail the hardware infrastructure and software architecture employed in our real-world experiments. The experimental setup centers on the Unitree G1 humanoid robot \cite{unitree2024g1}, supported by an external workstation for high-level inference and HTC Vive Ultimate Trackers for global localization \cite{HTC2024VIVEUltimateTracker}. Below, we describe the custom end-effector design, the perception setup, and the hierarchical control system.

\subsection{Gripper Design}
To endow the Unitree G1 with manipulation capabilities while minimizing distal mass, we developed a custom gripper adapted from the UMI hardware interface \cite{chi2024universal}. We replaced the original spring-trigger mechanism with a direct-drive transmission inspired by the Wild LMA design. Specifically, we utilized the robot's existing wrist yaw motor to actuate the gripper, engineering a custom master gear that mates precisely with the motor's spline. This design allows us to remove the stock rubber hands and clamping mechanisms, securing the new mount via the original screw interfaces. While the transmission system was redesigned to enable direct actuation, the finger geometry and camera mount remain identical to the original UMI gripper to ensure compatibility with our training data.

\subsection{Camera Setup}
Visual observations are captured using two GoPro Hero 10 cameras \cite{GoProHero10} mounted on the grippers. Following the UMI hardware stack \cite{chi2024universal}, we utilize a GoPro Media Mod to output HDMI signals, which are converted to a low-latency USB 3.0 UVC interface via an Elgato HD60X capture card. These streams are transmitted to the workstation via USB, ensuring real-time observation updates.

\subsection{System Architecture}
As outlined in the main text, our framework comprises a hierarchical structure: a high-level manipulation policy and a low-level whole-body controller. Communication between the external workstation and the robot's onboard computer is established via a local wireless network using ZeroMQ.

\subsubsection{High-Level Policy}
The high-level policy runs on the workstation at 5\,Hz. It aggregates visual streams from the external cameras and proprioceptive data (received from the robot) to infer desired end-effector keypoint trajectories and gripper commands. These targets are published to the robot via the ZeroMQ interface.

\subsubsection{Low-Level Controller}
The low-level whole-body controller executes directly on the robot's onboard computer at 50\,Hz. Upon receiving the target keypoints and gripper commands, it computes the necessary joint position commands, which are executed by the robot's built-in PD controller. To support precise tracking, we attach one HTC Vive Ultimate Tracker to the robot's pelvis for global localization and place a second tracker on the ground to serve as a static $z=0$ reference frame. Additional proprioceptive states, such as joint positions and IMU readings, are accessed directly from onboard sensors and streamed back to the high-level policy.

\section{Additional Experiments}
\label{sec:appendix_additional_experiments}
\subsection{Adaptive End-Effector Reward Ablation}
\label{sec:appendix_adaptive_ee_ablation}
We evaluate a naive alternative to the adaptive end-effector (EE) reward: enforcing uniformly tight EE tracking throughout the motion. We choose the \textbf{squat-and-retrieve} task because it requires accurate near-ground EE control as well as whole-body stability during a deep squat.

\textbf{\uline{Remove adaptive end-effector reward.}} For this baseline, we replace the adaptive end-effector (EE) reward with a fixed formulation:

\begin{equation*}
     r_{\text{EE}} = \sum_{\chi \in \{\mathbf{p}, \theta\}} r(\bar{e}_{\chi}^{\text{EE}}, \sigma_{\chi}^{\text{fixed}}),
\end{equation*}
where we set $\sigma_{\chi}^{\text{fixed}}$ to the minimum tolerance in our adaptive reward ($\sigma_{\mathbf{p}}^{\text{fixed}} = 0.01\,\text{m}$, $\sigma_{\theta}^{\text{fixed}} = 5^\circ$), and fix the weight to $\omega_{\text{EE}}^{\text{fixed}} = \omega_{\text{EE}}^{\max} = 0.5$. This configuration enforces a uniformly tight EE constraint throughout the motion, regardless of motion phase or modality. Under this setting, the success rate drops from \textbf{17/20 = 85\%} to \textbf{5/10 = 50\%}. Typical failures occur during the deep-squat phase: the humanoid either loses balance and falls, or struggles to settle into an unnatural, marginally stable pose that prevents subsequent manipulation. This degradation highlights the critical role of the adaptive EE reward. By dynamically shaping the reward landscape according to motion modalities, our method balances whole-body stability and manipulation precision throughout training. In contrast, permanently prioritizing EE precision forces the policy to neglect essential whole-body dynamics, compromising stability during whole-body dynamic phases like squatting.

\subsection{Additional TWIST2 Comparisons}
\label{sec:appendix_twist2_additional_comparisons}
\begin{figure}[H]
    \centering
    \captionsetup[subfigure]{skip=1pt}
    \begin{subfigure}[t]{0.42\linewidth}
        \centering
        \includegraphics[width=\linewidth]{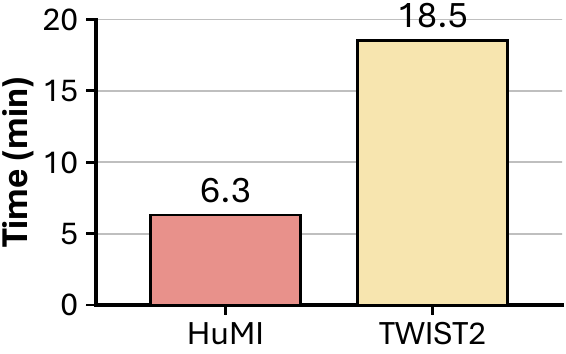}
    \end{subfigure}
    \hfill
    \begin{subfigure}[t]{0.54\linewidth}
        \centering
        \includegraphics[width=\linewidth]{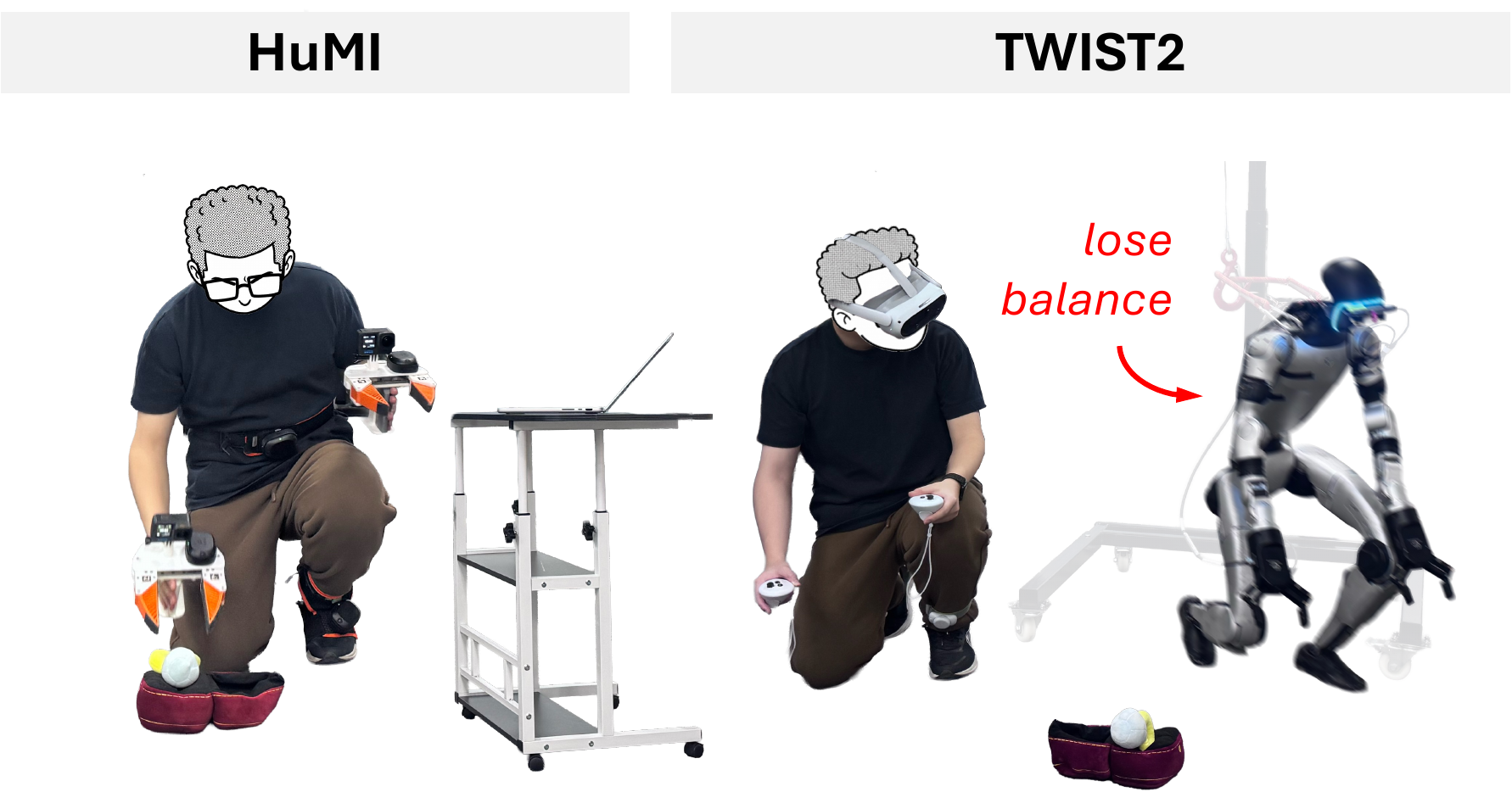}
    \end{subfigure}
    \caption{\textbf{Additional TWIST2 comparisons.} Left: time required to collect 100 demonstrations for bimanual pick-and-place. Right: illustration of the single-knee pick-up task.}
    \label{fig:appendix_twist2_additional}
\end{figure}

We further compare HuMI with TWIST2~\cite{ze2025twist2} on two additional settings. First, we evaluate the bimanual pick-and-place task reported in the TWIST2 paper. Under the same target of collecting 100 demonstrations, HuMI completes the collection in \textbf{6.3\,min}, whereas TWIST2 reports \textbf{18.5\,min} (left of Fig.~\ref{fig:appendix_twist2_additional}). This result is consistent with the throughput comparison in Fig.~\ref{fig:twist_comparison}: teleoperation-free data collection substantially reduces the time required to obtain usable demonstrations.

Second, we evaluate the \textbf{single-knee pick-up} task (right of Fig.~\ref{fig:appendix_twist2_additional}). In a 15\,min session, HuMI collects \textbf{50} usable demonstrations. Under the same setting, TWIST2 was unable to complete usable demonstrations for this task, as its controller could not reliably support the single-knee transition without balance failures. This reflects a controller-side constraint in teleoperation-based data collection: collected motions are limited by the current controller's capabilities.

\section{Generalization Experiments Details}
\label{sec:appendix_generalization_experiments}

\begin{figure}[h]
    \centering
    \includegraphics[width=1\linewidth]{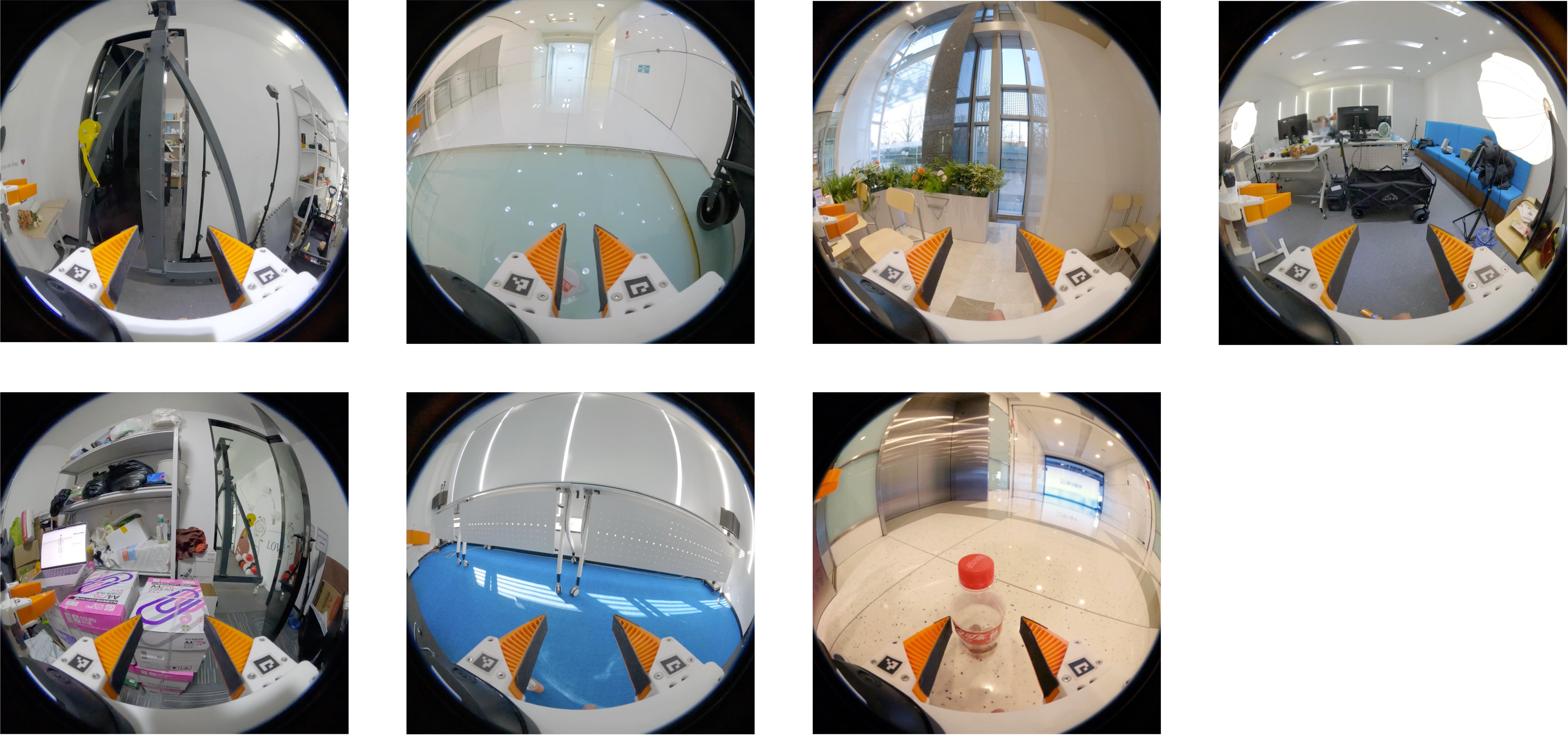}
    \caption{\textbf{Training environments}.}
    \label{fig:appendix_training_environment}
\end{figure}

\begin{figure}[h]
    \centering
    \includegraphics[width=1\linewidth]{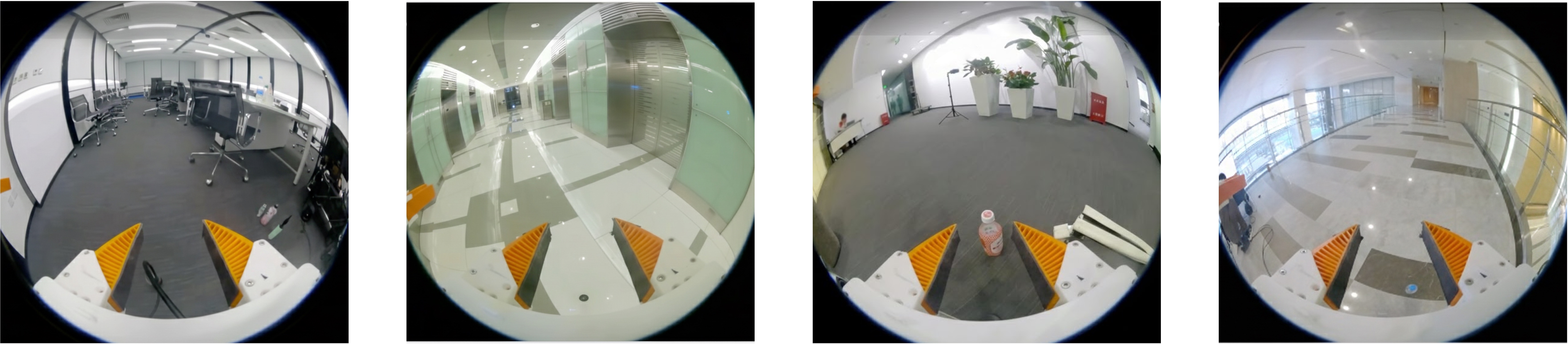}
    \caption{\textbf{Testing environments}.}
    \label{fig:appendix_testing_environment}
\end{figure}
\textbf{Environment details.} Fig.~\ref{fig:appendix_training_environment} visualizes the seven training environments in Sec.~\ref{sec:exp_generalization}. We collected 50 demonstrations in each environment, resulting in a total of 350 training trajectories. Fig.~\ref{fig:appendix_testing_environment} visualizes the four held-out testing environments; we conducted five real-world rollouts in each environment, for 20 rollouts in total.

\begin{figure}[h]
    \centering
    \begin{minipage}[t]{0.51\linewidth}
        \centering
        \includegraphics[width=\linewidth]{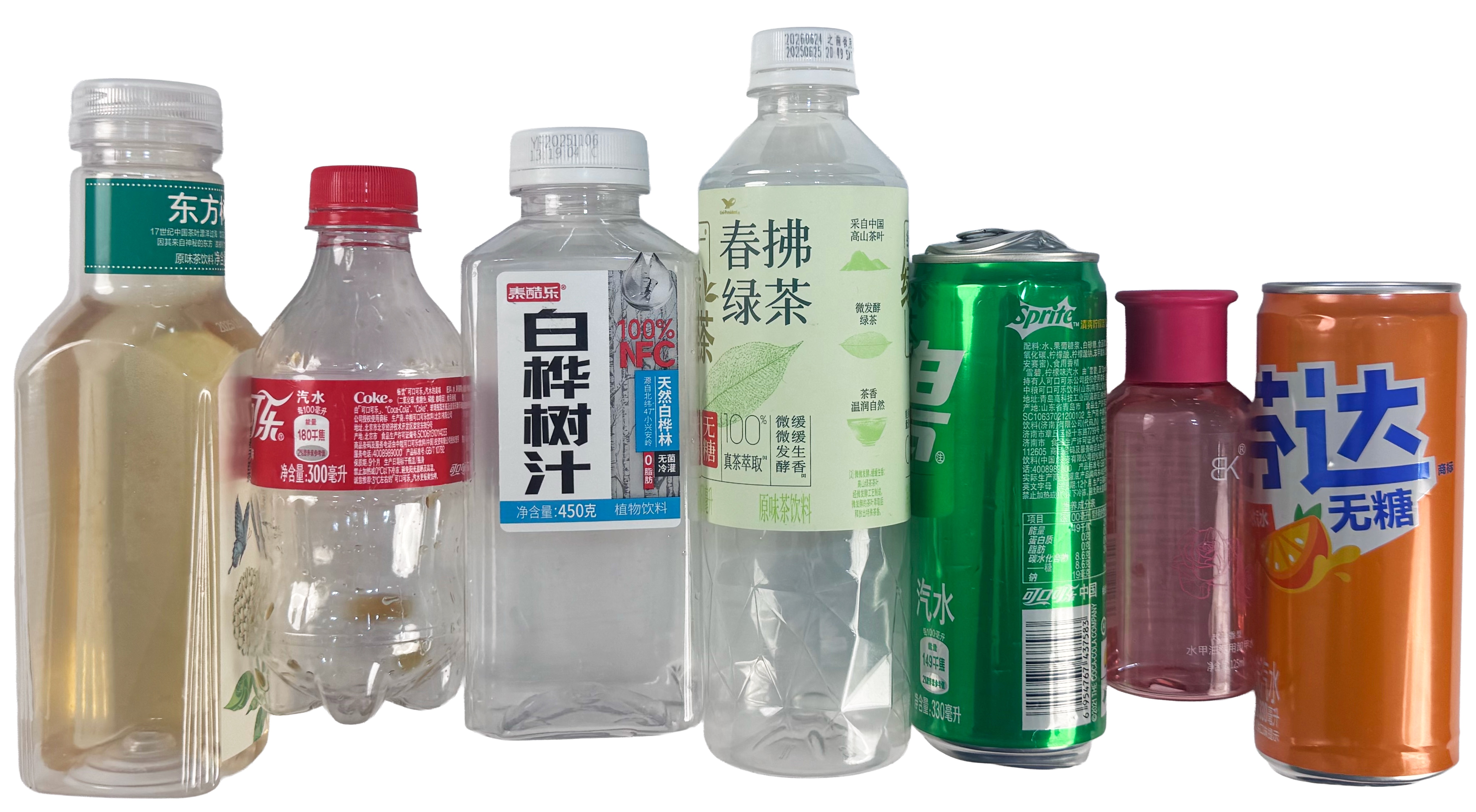}
        \caption{\textbf{Training objects}.}
        \label{fig:appendix_training_object}
    \end{minipage}
    \hfill
    \begin{minipage}[t]{0.35\linewidth}
        \centering
        \includegraphics[width=\linewidth]{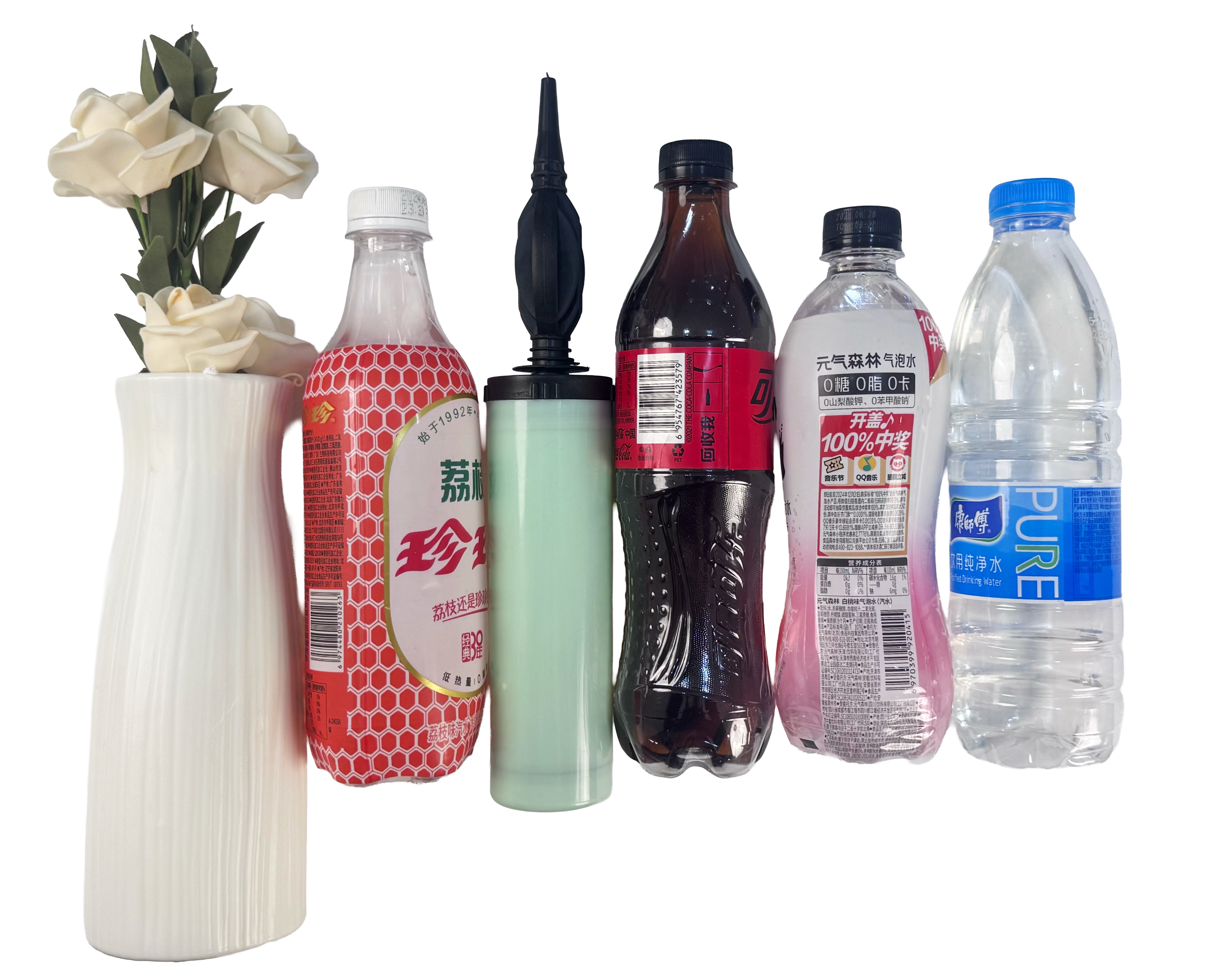}
        \caption{\textbf{Testing objects}.}
        \label{fig:appendix_testing_object}
    \end{minipage}
\end{figure}

\textbf{Object details.} Fig.~\ref{fig:appendix_training_object} visualizes the seven training objects used for our policy in Sec.~\ref{sec:exp_generalization}. Fig.~\ref{fig:appendix_testing_object} visualizes the six held-out testing objects, which are used in all real-world rollouts.

\section{Low-Level Controller Training Details}
\label{sec:appendix_training_details}

\textbf{Observation.} We train the low-level controller in a teacher–student framework. Similar to previous works~\cite{he2024omnih2o,he2025hover,li2025clone,ze2025twist}, we first train a teacher tracker with access to privileged states and full-body reference commands. We then distill this teacher into a student policy that operates only on real-world states and keypoint trajectories aligned with the high-level policy, using the DAgger algorithm.

Specifically, at time step $t$, the teacher's observation is
$O^{\text{tea}}_{t}=\left[s^{\text{tea}}_{t},a^{\text{tea}}_{t-1},c^{\text{tea}}_{t} \right]$,
where the state
$s^{\text{tea}}_{t}=\left[\mathbf{q}_{t}, \dot{\mathbf{q}}_{t}, \omega_{t}, g_{t} \right]$
includes the full-body joint positions and velocity $\mathbf{q}_{t}$, $\dot{\mathbf{q}}_{t}$, the base angular velocity $\omega_{t}$, and the base gravity vector $g_{t}$ projected into the body frame; $a^{\text{tea}}_{t-1}$ is the previous action. The whole-body command
$c^{\text{tea}}_{t}=\left[\mathbf{q}_{t}^{\text{ref}}, \dot{\mathbf{q}}_{t}^{\text{ref}}, \mathbf{p}_{t}^{\text{ref}}, \theta_{t}^{\text{ref}}, \mathbf{p}_{t}^{\text{ref}}-\mathbf{p}_{t}, \theta_{t}^{\text{ref}}\ominus\theta_{t}\right]$
contains the reference joint positions and velocities $\mathbf{q}_{t}^{\text{ref}}, \dot{\mathbf{q}}_{t}^{\text{ref}}$, as well as the reference link positions and orientations $\mathbf{p}_{t}^{\text{ref}}, \theta_{t}^{\text{ref}}$ together with their deviations from the current link poses $\mathbf{p}_{t}, \theta_{t}$.

For the student policy, the observation is
$O^{\text{stu}}_{t}=\left[s^{\text{stu}}_{t-25:t},a^{\text{stu}}_{t-26:t-1},c^{\text{stu}}_{t} \right]$,
where $s^{\text{stu}}_{t}=\left[\mathbf{q}_{t}, v_{t}, \omega_{t}, g_{t} \right]$ denotes the same real-world state features as above, and we include a history of 25 steps of states and past actions to provide temporal context. The student command strictly aligns with the high-level policy output,
$c^{\text{stu}}_{t}=\left[c^{\text{EE}}_{\mathcal{T}_t},c^{\text{blind}}_{\mathcal{T}_t} \right]$,
where each component contains a list of 10 waypoints sampled over the next $2\,\text{s}$:
\[
\mathcal{T}_t=\{\,t_{k}=t+k\cdot\Delta t\,|\,k=1,\dots,10\},
\quad
\Delta t=\left\lfloor\frac{2}{10\,\delta t}\right\rfloor,
\]
and the control time step is $\delta t=1/50\,\text{s}$.

The end-effector command at time step $t_k$ is
$
c^{\text{EE}}_{t_{k}}=\left[\mathbf{p}^{\text{ref}}_{t_{k}}, \theta^{\text{ref}}_{t_{k}}, \mathbf{p}^{\text{ref}}_{t_{k}}-\mathbf{p}_{t_{k}},  \theta^{\text{ref}}_{t_{k}}\ominus\theta_{t_{k}}\right],
$
which includes the localized reference end-effector position and orientation  $\mathbf{p}^{\text{ref}}_{t_{k}}, \theta^{\text{ref}}_{t_{k}}$, together with the position and orientation deltas, $\mathbf{p}^{\text{ref}}_{t_{k}}-\mathbf{p}_{t_{k}}$ and $\theta^{\text{ref}}_{t_{k}}\ominus\theta_{t_{k}}$. For ``blind'' points such as the pelvis or feet, we instead use relative displacements with respect to the current reference pose:
$
c^{\text{blind}}_{t_{k}}=\left[\mathbf{p}^{\text{ref}}_{t_{k}}-\mathbf{p}^{\text{ref}}_{t},  \theta^{\text{ref}}_{t_{k}}\ominus\theta^{\text{ref}}_{t}\right],
$
which keeps the command \emph{relative within each action chunk}, independent of the absolute position.

\textbf{Reward design.}
Following prior works~\cite{liao2025beyondmimic, luo2025sonic}, we decompose the low-level reward into a tracking term $r_{\text{tracking}}$ and a regularization term $r_{\text{penalty}}$.
Table~\ref{tab:rewards} summarizes all reward terms for training.
\begin{table}[!t]
    \centering
    \caption{\textbf{Rewards used in low-level controller training.}}
    \label{tab:rewards}
    \vspace{2pt}
    \small
    \setlength{\tabcolsep}{3pt}
    \renewcommand{\arraystretch}{1.12}

    \resizebox{\linewidth}{!}{%
    \begin{tabular}{@{}l l c@{}}
        \toprule
        \textbf{Reward Term} & \textbf{Equation} & \textbf{Weight} \\
        \midrule
        \rowcolor{humiTableGray}
        \multicolumn{3}{@{}l}{\textit{Tracking Rewards}} \\
        Whole-body position tracking &
        $\exp(-\|\mathbf{p}^{\text{ref}}-\mathbf{p}\|_2^2/0.3^2)$
        &1.0 \\
        Whole-body rotation tracking &
        $\exp(-\|\mathbf{\theta}^{\text{ref}}\ominus\mathbf{\theta}\|_2^2/0.4^2)$
        &1.0 \\
        Whole-body linear velocity tracking &
        $\exp(-\|\mathbf{v}^{\text{ref}}-\mathbf{v}\|_2^2/1.0^2)$
        &1.0 \\
        Whole-body angular velocity tracking &
        $\exp(-\|\mathbf{\omega}^{\text{ref}}-\mathbf{\omega}\|_2^2/\pi^2)$
        &1.0 \\
        Adaptive EE position tracking &
        $\mathbb{I}(\|v^{\text{ref}}_{\text{base}}\| < 0.02\,\mathrm{m/s}) \cdot \exp(-\|\mathbf{p}_{\text{EE}}^{\text{ref}}-\mathbf{p}_{\text{EE}}\|_2^2/\sigma_\mathbf{p}(v^{\text{ref}}_{\text{EE}})^2)$
        & $0.0 \rightarrow 0.5$ \\
        Adaptive EE rotation tracking &
        $\mathbb{I}(\|v^{\text{ref}}_{\text{base}}\| < 0.02\,\mathrm{m/s}) \cdot \exp(-\|\mathbf{\theta}_{\text{EE}}^{\text{ref}}\ominus\mathbf{\theta}_{\text{EE}}\|_2^2/\sigma_\theta(v^{\text{ref}}_{\text{EE}})^2)$
        & $0.0 \rightarrow 0.5$ \\
        \addlinespace[3pt]
        \rowcolor{humiTableGray}
        \multicolumn{3}{@{}l}{\textit{Regularization Penalty}} \\
        \addlinespace[1pt]
        Action rate &
        $\|a_t - a_{t-1}\|_2^2$ &
        $-5\times 10^{-2}$ \\
        Joint limits & $\sum\mathbb{I}(q_j \notin (q_{j}^{\min}, q_{j}^{\max}))$ & -10.0 \\
        Undesired contact &
        $\sum_{i \notin \{\text{ankles}, \text{knees}, \text{hips}\}}\mathbb{I}\!\left( \|\mathbf{F}_i\|_2 > 1.0\,\mathrm{N} \right)$ &
        -0.1 \\
        \bottomrule
    \end{tabular}
    }
\end{table}

\noindent\textbf{Domain randomization.} Table~\ref{tab:domain_rand} summarizes the domain randomizations used for low-level controller training, grouped into physical, velocity, reset, and speed randomization.
\begin{table}[!t]
    \centering
    \caption{\textbf{Domain randomization used in low-level controller training.}}
    \label{tab:domain_rand}
    \vspace{2pt}
    \small
    \setlength{\tabcolsep}{3pt}
    \renewcommand{\arraystretch}{1.12}

    \begin{tabular}{@{}l p{0.42\linewidth} p{0.28\linewidth}@{}}
        \toprule
        \textbf{Term} & \textbf{Description} & \textbf{Sampling Range} \\
        \midrule
        \rowcolor{humiTableGray}
        \multicolumn{3}{@{}l}{\textit{Physical randomization}} \\
        Static friction & randomize static friction of robot bodies & $\mu_{s} \sim \mathcal{U}[0.3, 1.6]$ \\
        Dynamic friction & randomize dynamic friction of robot bodies & $\mu_{d} \sim \mathcal{U}[0.3, 1.2]$ \\
        Restitution & randomize restitution of robot bodies & $e \sim \mathcal{U}[0.3, 1.2]$ \\
        Default joint positions & add offsets to default joint positions & $\mathbf{q}_0 \sim \mathbf{q}_0+\mathcal{U}[0.0, 0.5]$ \\
        CoM offsets & randomize the torso link center of mass & \begin{tabular}[t]{@{}l@{}}$\Delta x \sim \mathcal{U}[-0.025, 0.025]$,\\ $\Delta y,\Delta z \sim \mathcal{U}[-0.05, 0.05]$\end{tabular} \\
        End-effector mass & randomize the end-effector mass & $m \sim \mathcal{U}[0.75, 1.25]\cdot m$ \\
        \addlinespace[3pt]
        \rowcolor{humiTableGray}
        \multicolumn{3}{@{}l}{\textit{Velocity perturbations}} \\
        Push robot & periodically push the robot every $\Delta t$ & \begin{tabular}[t]{@{}l@{}}$\Delta t \sim \mathcal{U}[4, 6]$,\\ $v_{\text{x,y}} \sim \mathcal{U}[-0.5, 0.5]$,\\ $\omega_{\text{yaw}} \sim \mathcal{U}[-0.78, 0.78]$\end{tabular} \\
        \addlinespace[3pt]
        \rowcolor{humiTableGray}
        \multicolumn{3}{@{}l}{\textit{Reset perturbations}} \\
        Base position & perturb base position at the reset time & \begin{tabular}[t]{@{}l@{}}$\Delta p_{\text{x,y}} \sim \mathcal{U}[-0.05, 0.05]$,\\ $\Delta p_{\text{z}} \sim \mathcal{U}[0.0, 0.05]$\end{tabular} \\
        Base orientation & perturb base orientation at the reset time & \begin{tabular}[t]{@{}l@{}}$\Delta \omega_{\text{pitch},\text{roll}} \sim \mathcal{U}[-0.1, 0.1]$,\\ $\Delta \omega_{\text{yaw}} \sim \mathcal{U}[-0.2, 0.2]$\end{tabular} \\
        Base linear velocity & perturb base linear velocity at the reset time & \begin{tabular}[t]{@{}l@{}}$\Delta v_{\text{x,y}} \sim \mathcal{U}[-0.05, 0.05]$,\\ $\Delta v_{\text{z}} \sim \mathcal{U}[-0.2, 0.2]$\end{tabular} \\
        Base rotation velocity & perturb base rotation velocity at the reset time & \begin{tabular}[t]{@{}l@{}}$\Delta \omega_{\text{roll,pitch}} \sim \mathcal{U}[-0.52, 0.52]$,\\ $\Delta \omega_{\text{yaw}} \sim \mathcal{U}[-0.78, 0.78]$\end{tabular} \\
        Joint position & perturb joint positions at the reset time & $\Delta q \sim \mathcal{U}[-0.1, 0.1]$ \\
        Reset pose shift & shift reset pose around $t_{\text{reset}}$ & $t_{\text{reset}}\sim t_{\text{reset}}+\mathcal{U}[-0.05, 0.05]$ \\
        \addlinespace[3pt]
        \rowcolor{humiTableGray}
        \multicolumn{3}{@{}l}{\textit{Speed randomization}} \\
        Variable-speed augmentation & sample a play speed every 0.02\,s & $s\sim \mathcal{N}_{[0.25,\,1.25]}(1.0,\sigma^2)$ \\
        \bottomrule
    \end{tabular}
\end{table}

For the adaptive EE tracking rewards, we compute the tolerance parameter $\sigma_{\chi}$ by linearly interpolating between $\left[\sigma_\chi^{\min}, \sigma_\chi^{\max}\right]$ as a function of the commanded end-effector velocity $v^{\text{ref}}_{\text{EE}}$:
\begin{equation*}
\begin{aligned}
    &\sigma_{\chi}\bigl(v^{\text{ref}}_{\text{EE}}\bigr)
    = \operatorname{clip}\!\left(f_{\text{interp}}(v^{\text{ref}}_{\text{EE}})
   ,
    \, \sigma_{\chi}^{\min}, \, \sigma_{\chi}^{\max}
    \right),\\
    &f_{\text{interp}}(v^{\text{ref}}_{\text{EE}})=\frac{v^{\text{ref}}_{\text{EE}} - v^{\min}}{v^{\max}-v^{\min}}(\sigma_{\chi}^{\max} - \sigma_{\chi}^{\min}) + \sigma_{\chi}^{\min}.
\end{aligned}
\end{equation*}

For the adaptive end-effector position tracking rewards, we set $\sigma_\mathbf{p}^{\min}=0.01\,\mathrm{m}$ and $\sigma_\mathbf{p}^{\max}=0.1\,\mathrm{m}$. For the adaptive end-effector rotation tracking rewards, we set $\sigma_\theta^{\min}=5^\circ$ and $\sigma_\theta^{\max}=20^\circ$. For both rewards, we set $v^{\max}=0.1\,\mathrm{m/s}$ and $v^{\min}=0.05\,\mathrm{m/s}$. Their reward weights are linearly increased from $0.0$ to $0.5$, and $\sigma_\mathbf{p}^{\min}$ is decreased from $0.1\,\mathrm{m}$ to $0.01\,\mathrm{m}$ over training steps $10{,}000$ to $15{,}000$.

The \emph{reset pose shift} targets the mismatch between the command and the robot state at deployment. During training, the low-level policy tracks replayed commands, while at test time the high-level policy issues online commands that may be temporally misaligned or abruptly corrected. To expose the controller to such offsets, we sample a reset time $t_{\text{reset}}$ along the demonstration and initialize the robot to the pose at $t_{\text{reset}} + \mathcal{U}[-0.05, 0.05]$, while keeping the original command trajectory.

For \emph{variable-speed augmentation}, every $0.02$\,s we sample a speed scaling factor $s \sim \mathcal{N}_{[0.25,\,1.25]}(1.0, \sigma^2)$ and use it to scale the reference time, where $\mathcal{N}_{[0.25,\,1.25]}$ denotes a normal distribution truncated to $[0.25,\,1.25]$. The standard deviation $\sigma$ is linearly increased from $10^{-4}$ to $1.0$ between training steps $10{,}000$ and $15{,}000$, after which it is kept at $\sigma=1.0$.

\section{High-Level Policy Training Details}
\label{sec:appendix_highlevel_training_details}
We employ Diffusion Policy \cite{chi2025diffusion} as the high-level backbone. Notably, this component is designed for modularity; alternative architectures, such as ACT \cite{zhao2023learning}, can be seamlessly substituted.

\subsection{Hyperparameters}
We use the same training hyperparameters for all tasks, as summarized in Table~\ref{tab:dp_params}.
\begin{table}[!h]
    \centering
    \caption{\textbf{Hyperparameters for the high-level diffusion policy.}}
    \label{tab:dp_params}
    \begin{tabular}{lr}
        \toprule
        \textbf{Hyperparameter} & \textbf{Value} \\
        \midrule
        Visual observation horizon & 1 \\
        Visual observation frequency & 20 Hz \\
        Proprioceptive observation horizon & 3 \\
        Proprioceptive observation frequency & 20 Hz \\
        Action horizon & 48 \\
        Action frequency & 20 Hz \\
        Execution-to-data speed ratio & $1\times$ \\
        Image resolution ($N_{\text{cam}} \times H \times W$) & $2 \times 224 \times 224$ \\
        Vision backbone &  \texttt{vit\_base\_patch14\_dinov2.lvd142m}\\
        Diffusion Policy learning rate & $3\text{e-}4$ \\
        Vision backbone learning rate & $3\text{e-}5$ \\
        Epochs & 200 \\
        Batch size & 256 \\
        Training loss & Flow Matching \\
        Inference denoising steps & 10 \\
        \bottomrule
    \end{tabular}
\end{table}

\subsection{Observation and Action Spaces}
The high-level policy conditions on both visual and proprioceptive observations. Visual inputs comprise RGB images captured by two gripper-mounted cameras, each with a resolution of $224 \times 224$ pixels. Proprioceptive observations consist of the robot's lower-body joint angles. The action space defines the desired positions and rotations of the keypoints, along with gripper widths. We adopt the same action parameterization as UMI \cite{chi2024universal}.

\subsection{Training Data}
For each capability task, including single-knee pick-up, sword-unsheathing, toy-tossing, and walking-to-clean, we collect 100 demonstrations in a single environment. For the squat-and-retrieve task, we collect 350 demonstrations across seven distinct environments ($7\times50$).

\end{document}